\documentclass[10pt,twocolumn]{IEEEtran}
\usepackage{graphicx}           
\usepackage{amsmath,amsfonts}
\usepackage{times,epsfig}
\usepackage{psfrag}
\usepackage{subfigure}
\usepackage{stfloats}
\usepackage{amssymb}
\usepackage{multirow}
\usepackage[justification=centering]{caption}
\usepackage{cite}
\usepackage{CJK}
\usepackage{url}
\usepackage{caption}
\usepackage{mathrsfs}
\usepackage{bm}
\usepackage{array}
\usepackage{algorithmic}
\usepackage[lined,ruled,linesnumbered]{algorithm2e}

\newcounter{mytempeqncnt}

\begin{document}
\pagenumbering{arabic}

\title{FedParking: A Federated Learning based Parking Space Estimation with Parked Vehicle assisted Edge Computing}

\author{Xumin Huang, Peichun Li, \emph{Student Member, IEEE}, Rong Yu, \emph{Member, IEEE}, \\ Yuan Wu, \emph{Senior Member, IEEE}, Kan Xie, and Shengli Xie, \emph{Fellow, IEEE}
}
\maketitle
\thispagestyle{empty}

\begin{abstract}
As a distributed learning approach, federated learning trains a shared learning model over distributed datasets while preserving the training data privacy. We extend the application of federated learning to parking management and introduce FedParking in which Parking Lot Operators (PLOs) collaborate to train a long short-term memory model for parking space estimation without exchanging the raw data. Furthermore, we investigate the management of Parked Vehicle assisted Edge Computing (PVEC) by FedParking. In PVEC, different PLOs recruit PVs as edge computing nodes for offloading services through an incentive mechanism, which is designed according to the computation demand and parking capacity constraints derived from FedParking. We formulate the interactions among the PLOs and vehicles as a multi-lead multi-follower Stackelberg game. Considering the dynamic arrivals of the vehicles and time-varying parking capacity constraints, we present a multi-agent deep reinforcement learning approach to gradually reach the Stackelberg equilibrium in a distributed yet privacy-preserving manner. Finally, numerical results are provided to demonstrate the effectiveness and efficiency of our scheme.
\end{abstract}


\section{Introduction}
Nowadays, federated learning has been envisioned as a distributed learning framework in which devices cooperate to build and update a globally shared learning model by supporting training the global model over distributed datasets \cite{mcmahan2017, yang2019}.  Federated learning enables each device as a data owner to locally train the global model with individually collected data. This approach exchanges model parameters instead of the actual training data and preserves data privacy of the devices. Due to the significant privacy-friendly characteristic, federated learning is applied to diverse domains, e.g.,  mobile keyboard prediction \cite{hard2018}, wearable activity recognition \cite{9076082} and content caching placement \cite{9060924}. The promising approach is also integrated into vehicular networks to forecast traffic information such as traffic flow \cite{9082655} and traffic speed \cite{9340313} while guaranteeing reliable data privacy preservation during the forecast. Motivated by the current works, we extend the application of federated learning to parking management in smart cities, and develop a federated learning based parking space estimation scheme named by FedParking. 

Fedparking  allows multiple Parking Lot Operators (PLOs) to collaboratively learn a  Long Short-Term Memory (LSTM) model for parking space estimation while keeping their training data secure and private. On one hand,  a PLO should predict the number of free parking spaces in real time for traffic management at a parking lot. To achieve parking space estimation,  researchers have presented a variety of applicable methods. Since prediction of parking space availability including parking occupancy and duration of occupancy can be regarded as a time-series problem,  time series prediction is utilized as a promising solution to make full use of prior knowledge from historical data and obtain accurate prediction results. Furthermore, the time-series prediction problem on parking space estimation is characterized by the long-term trend and cyclical fluctuations, and LSTM as a powerful variant of the typical recurrent neural network is particularly suitable to tackle it \cite{gers2002learning}. Thus, we adopt a typical LSTM model as the global model in Fedparking, which is shared to each PLO to achieve an accurate estimation of the parking spaces.

On the other hand, the global model can simply feed off the training data extracted from parking space information of all the PLOs. In this regard, a naive learning approach usually requires each PLO to upload the training data and utilizes the collected data for centralized training. However, the traditional learning approach puts individual privacy of vehicles at risk, since the parking space information records basic profiles of daily parking requests such as arrival and departure time of the vehicles. The parking space information also records time-varying parking price and occupancy of the parking lots. For commercial purposes,  PLOs would not like to reveal the parking space information to the others to prevent malicious entities which infer the operating mode of the parking lots. This means that a secure and distributed learning approach is necessitated such that different PLOs collaborate to train the global model without potential security vulnerabilities.  We pay attention to the recently emerged federated learning that does not rely on sharing training data among the clients and propose FedParking. In FedParking, different PLOs collaborate to train the global model by the local datasets and only feedback the updated model parameters for parameter aggregation.  To summarize, we are motivated to study an integration of the federated learning and LSTM to facilitate the FedParking design with efficiency and security guarantee. 

Building upon the FedParking for predicting the parking spaces at different PLOs, we devise an incentive mechanism for Parked Vehicle assisted Edge Computing (PVEC).  As an application of mobile edge computing in Internet of vehicles, PVEC schedules stochastic idle communication and computing resources from Parked Vehicles (PVs). It can significantly establish an available and cost-effective computing resource pool, which alleviates computational workloads and augments system capacity \cite{9090367}.  A PLO acts as an offloading service provider to receive computation-intensive but not delay-sensitive tasks from nearby offloading users, afterwards stimulates vehicles to enter the parking spaces and utilize the on-board resources of the PVs to accomplish the given workloads. To realize the goal,  an appropriate incentive mechanism is necessitated for the PLO to provide vehicles with monetary rewards according to the parking capacity and computation demand. Through FedParking, we can estimate the number of available parking spaces of every parking lot which leads to the corresponding parking capacity constraint for each PLO. 

After that, we  formulate the strategic interactions among multiple PLOs and vehicles as a two-stage and non-cooperative game, namely, a multi-leader multi-follower Stackelberg game. A PLO is a game leader that provides monetary rewards to attract the vehicles to enter the parking lot and share the idle computing resources in the first stage. In turn, the vehicles are the game followers which act in response to the PLOs' strategies. The Stackelberg game aims to improve the utilities of the PLOs while simultaneously guaranteeing utility maximization for each vehicle. By theoretical analysis, we prove the existence and uniqueness of the Stackelberg equilibrium while neglecting the parking capacity constraints. Furthermore, we apply Deep Reinforcement Learning (DRL) as a solution methodology to reach the Stackelberg equilibrium in practical scenarios where both arrivals of the vehicles and parking capacity constraints of the PLOs are varying over time. In our DRL approach,  each PLO is regarded as an agent to learn a near-optimal decision by referring to the historical strategies of the others. The approach depends on iterative interactions among the players and supports distributed decision making instead of holding any prior knowledge of all the players in advance.

The main contributions of the paper can be summarized as follows.
\begin{itemize}
	\item We adopt federated learning and LSTM to promote the collaboration among the PLOs in parking space estimation. In FedParking, each PLO trains a shared LSTM model to forecast available parking spaces  by the local data while preserving the training data privacy. Technique details including system model, network entities, workflow and LSTM model design are provided.

	\item We study the incentive mechanism design of PVEC by FedParking, which puts forward a parking capacity constraint for each PLO. A PLO needs to recruit PVs to complete offloading workloads and an appropriate incentive mechanism is necessitated to maximize the expected profit and simultaneously take into account the parking capacity constraint. After introducing a Stackelberg game model among the PLOs and vehicles, we provide the equilibrium analysis to validate the existence and uniqueness of Stackelberg equilibrium by studying the sub-game perfect equilibrium in each stage. 
	
	\item We present a DRL based incentive mechanism for the PLOs since the closed-form Stackelberg equilibrium is difficult to acquire in the non-static environment.  A multi-agent DRL approach is introduced to gradually reach the Stackelberg equilibrium in a distributed yet privacy-preserving manner. Our approach avoids collecting the private information of the players of the Stackelberg game and achieves the almost same convergence performance as the centralized approach.
\end{itemize}

The rest of this paper is organized as follows. Section II presents the related work.
We introduce the detailed FedParking design based on federated learning and LSTM in Section III. Section IV studies the Stackelberg game based incentive mechanism. Section V provides the theoretical Stackelberg equilibrium analysis without considering the parking capacity constraints. In Section VI, we propose the DRL approach to gradually reach the Stackelberg equilibrium with the given parking capacity constraints. Numerical results are shown in Section VII. Finally, Section VIII concludes this paper and discusses the future direction.

\section{Related Work}
\subsection{Federated Learning for Vehicular Networks}
As a privacy-preserving learning approach, federated learning enables the collaborative training of a globally shared learning model without exchanging raw data among the clients.  Recently, the promising tool has been applied to vehicular networks for different purposes. For example, federated learning was combined with a gated recurrent unit neural network to encourage multiple organizations to cooperatively predict future traffic flow without privacy leakage \cite{9082655}. The distributed learning technique was exploited to provide robust privacy-preserving traffic speed forecasting and protection of topological information \cite{9340313}. To eliminate the vehicular data leakage, the authors in \cite{9105934} presented a federated learning  based collaborative data leakage detection scheme. Besides,  federated learning was adopted among mobile vehicles to improve vehicular communications with ultra reliability and low latency \cite{8917592}, and support AI applications such as image classification \cite{8964354} in vehicular edge computing environment. 

Motivated by the above works, we propose an integration of federated learning and LSTM to design  FedParking and achieve the PLOs' collaborative parking space estimation. According to a federated learning based framework, various PLOs cooperate with a parameter server to jointly train the LSTM model in an efficient and secure manner. Under the privacy restrictions, FedParking only collects model parameters updated from the PLOs and does not require sharing the datasets of the PLOs with the parameter server. Moreover, owing to the flexibility of the federated learning based framework, FedParking is computation efficient and also compatible with the other machine learning models in addition to the proposed LSTM model for parking space estimation.  
\begin{figure*}[t]
	\centering
	\includegraphics[width=0.9\textwidth]{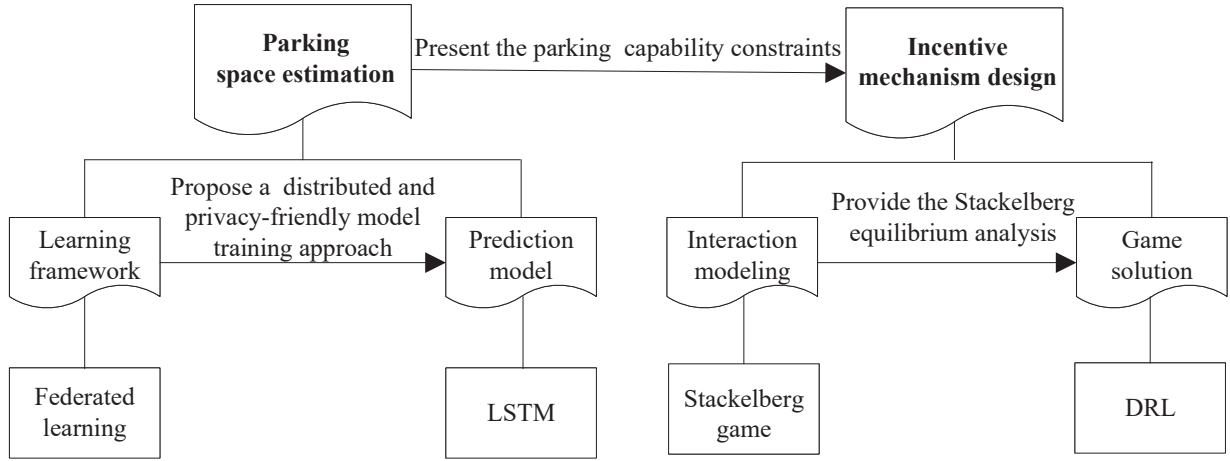}
	\caption{Methodology of FedParking.}
	\label{a1}
\end{figure*}
\subsection{Parked Vehicle Assisted Edge Computing}
Nowadays, many research efforts have been devoted to optimizing the performance of PVEC with different objectives. For example, a contract based incentive mechanism was designed to incentivize mobile and parked vehicles with different types to contribute their on-board resources for task processing \cite{8624349}. A contract was correlated with an offloading task and a contract item specifies the required number of computing resources and promising rewards for a vehicle type. The contract assignment between various tasks and vehicles was determined by using matching theory. The similar work was found in \cite{9372968}. Besides, mobile and  parked vehicles cooperate with nearby mobile edge computing servers for admission control \cite{8584057} and workload processing \cite{8931659, 9316937}. The vehicles also receive offloading tasks from unmanned aerial vehicles in smart city \cite{9364753} and help a local cloudlet cope with traffic packets and establish a distributed city-wide traffic management system \cite{8318667}. In addition to contract theory, game and auction theories have been applied for the incentive mechanism design in PVEC. A Stackelberg game was formulated to jointly employ a set of PVs and a mobile edge computing server for cooperative task processing \cite{9016397}. A multi-round multi-item parking reservation auction scheme was introduced to guide passing vehicles to specific parking spaces and make use of their processing capability to support proximal offloading services  \cite{zhang2019parking}. 

In spite the above studies, there still exist several important issues to be addressed for the success of PVEC. First, a PLO needs to dynamically predict the number of available parking spaces over time and adjust the incentive mechanism according to a parking capacity constraint and  computation demand. Second, previous works have not considered the competition effect among different PLOs, which provide  different monetary rewards for arriving vehicles. Due to selfishness in resource usage, the PLOs compete against each other to recruit the PVs and reserve their idle computing resources during the parking time. In turn,  parking choices of the vehicles will be influenced by the reward policies of the PLOs. The vehicles are rational to decide which parking lot to join and optimize the computing resources shared to the corresponding PLO.  Last but not the least, it is straightforward in most the previous works to collect private information of each PV and PLO for centralized decision making. This violates  privacy of the participants. Alternatively, a learning based incentive mechanism could be helpful for a PLO to seek a suboptimal solution  under incomplete information. 

Compared with the previous works, we propose a joint federated learning and DRL approach to promote the management of PVEC. A flowchart is illustrated in Fig.~\ref{a1} to present the comprehensive methodology of the proposed FedParking, which consists of two basic parts: parking space estimation and incentive mechanism design. Considering the parking management in smart cities, we extend the application of federated learning to parking space estimation and introduce FedParking. In the scheme, multiple PLOs collaborate to train a globally shared learning model to predict the number of free parking spaces in real time, without exchanging the raw data. Moreover, we utilize a LSTM model as the global model in FedParking to achieve an accurate estimation of the parking spaces. Building upon the above scheme for estimating the parking spaces at different PLOs, we further facilitate the incentive mechanism design for PVEC. According to the computing demand and parking capacity constraint, each PLO determines how to stimulate vehicles to enter the parking spaces and schedule idle  resources of the PVs to support local offloading services. We formulate the strategic interactions among multiple PLOs and vehicles as a multi-leader multi-follower Stackelberg game. The competition effect of different PLOs is considered. We temporarily neglect the parking capacity constraints, and provide the Stackelberg equilibrium analysis under the static conditions. We realize that the closed-form Stackelberg equilibrium is difficult to acquire in the practical scenarios where both arrivals of the vehicles and parking capacity constraints of the PLOs are varying over time. Thus, a DRL approach is applied, in which each PLO plays as an agent to learn a near-optimal decision, such that we can reach the Stackelberg equilibrium in a distributed manner.

\begin{figure}
	\centering
	\includegraphics[width=0.48\textwidth]{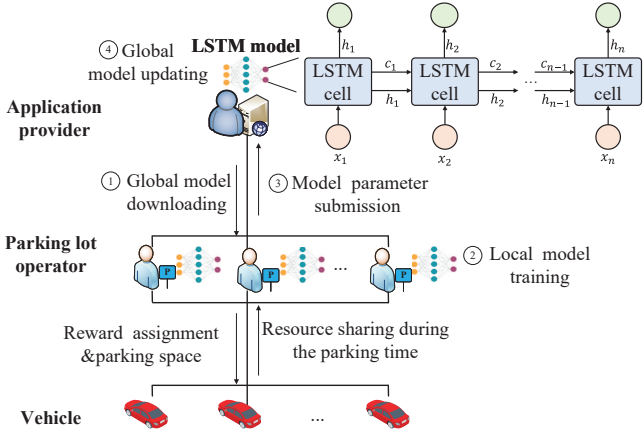}
	\caption{System design of FedParking.}
	\label{s0}
\end{figure}

\section{FedParking}
For traffic management at a parking lot, a PLO predicts the number of available parking spaces. To this end, a learning model is developed and utilized as a global model that is shared to multiple PLOs after iteratively training. To avoid the centralized data collection and centralized model training, we investigate the use of federated learning on parking space estimation, and design FedParking where many PLOs keep their raw data local but jointly train the global model by only sharing the updated model parameters with a parameter server. We first describe essential network entities of FedParking, and present the motivations and basic workflow of FedParking. Moreover, a LSTM model is adopted as the global model in FedParking, thereby improving the accuracy performance of parking space estimation.
\subsection{System Model and Network Entity}
In this work, we consider a system model shown in Fig.~\ref{s0}. Specifically, for parking demand, vehicles on the roads send their parking requests to an application provider of local parking services. The application provider notifies PLOs in the region to serve the vehicles.  Through PVEC, PLOs act as offloading service providers to attract the vehicles to enter their parking spaces by monetary rewards, and incentivize them to receive and complete given computation tasks during the parking time. We describe the essential network entities of FedParking as follows.


\begin{itemize}
	\item Parking Lot Operator: Each PLO manages a parking lot with limited parking spaces and specifies that an arriving vehicle occupies a parking space. The PLO adopts a learning model to estimate the number of free parking spaces and would like to participate in training the model by using the local data. In addition, when the vehicles are parked in a parking lot, their under-utilized on-board resources can be exploited to provision offloading services. Thus, there is great potential for tackling the computation demand of nearby users and service providers by offloading computation-intensive parts of mobile applications to the parking lot. Consequently, the PLO  acts as an employer to reserve computing resources from the PVs to complete the offloading workloads. Here, an incentive mechanism is required to provide economic compensations to the PVs.

	\item Parked Vehicle: A vehicle submits a parking request for parking space reservation. The parking request contains auxiliary information of the parking service, such as reservation time and original preferences over different parking lots.  By knowing current parking lots with their reward policies and workload requirements, the vehicle decides whether to enter a parking lot and share the computing resources to the PLO. For the vehicle, monetary rewards could compensate for parking fee and mainly influence the parking choice. After promising to accept the offloading workloads, the vehicle in the parking lot will receive input data from the PLO through a nearest roadside unit. The vehicle virtualizes the on-board computing resources by the containerization technology  to create the task instances and improve the task processing efficiency \cite{9090367}.
	
	\item Application Provider: The application provider is a totally trusted coordinator of the parking services in FedParking. It develops a client for vehicles that enables users with entrance to submit the parking requests and choose a suitable parking lot.  A client is also developed for PLOs to receive the parking requests and publish their reward policies. With the help of the clients,  a legitimate two-way communication channel between the vehicles and PLOs is built.
\end{itemize}

\subsection{Federated Learning for FedParking}
Learning-based time series prediction model is envisioned as a promising solution to tackle the problems of parking space estimation. The development of a time series prediction model inherits the methodology from traditional learning tasks, including data processing, model building, model training and inference.  In \cite{stolfi2017predicting}, the authors have developed a  data cleaning algorithm and trained various time series models for different parking lots. Given a training dataset with $M$ time series ${\mathcal D}=\{{\mathcal X}, {\mathcal Y}\}$,  we regard ${\mathcal X}=\{X_1, X_2, \cdots, X_M\}$ as the inputs and  ${\mathcal Y}=\{Y_1, Y_2, \cdots, Y_M\}$ as the corresponding ground truth predictions. Let $F_\Theta$ represent the time series model parameterized by weight $\Theta$. To train the time series prediction model, we minimize the Mean Square Error (MSE) between the estimation and the ground truth, i.e.,
\begin{equation} \label{loss_func}
\mathop {\min }\limits_\Theta  {\mathcal L}(\Theta , {\cal D}) \leftrightarrow \mathop {\min }\limits_\Theta  \frac{1}{M}\sum\limits_m {{{\left( {{F_\Theta }({X_m}) - {Y_m}} \right)}^2}}.
\end{equation}
In this paper, we consider a feasible scenario where multiple PLOs train and adopt a shared time series prediction model to forecast their available parking spaces .

With the privacy considerations, we introduce FedParking to facilitate the collaboration among the PLOs in parking space estimation. In FedParking,  there exist several communication rounds for the PLOs. In each communication round, each PLO firstly trains a  global  model with local data, and  then submits the updated model parameters to aggregate a new global model. According to the concept of federated learning, knowledge of the local models can be shared without revealing the training data to each other. The distributed learning approach finally preserves the data privacy of the participants.

We consider that there exist $I$ vehicles with parking demand and $J$ parking lots. Each vehicle $i, 1\le i\le I$ can choose whether to enter a parking lot $j, 1 \le j \le J$. Each PLO has obtained the historical data of parking space information. Let $\mathcal {D}_j$  represent a whole dataset of training data samples of PLO $j$, and $\Theta _0$ indicate the weight of the global  model. Under the federated learning based framework, the goal of FedParking is expressed by
\begin{equation}\label{fed_obj}
\mathop {\min }\limits_{{\Theta _0}} \frac{{\left\| {{{\mathcal D}_j}} \right\|}}{{\sum\limits_{1\le j \le J} {\left\| {{{\mathcal D}_j}} \right\|} }}\sum\limits_{1\le j \le J} {{\mathcal L}({\Theta _0},{{\mathcal D}_j})},
\end{equation}
where ${\left\| {{{\mathcal D}_j}} \right\|}$ denotes the training data size of PLO $j$, and $\mathcal L(\cdot)$ is the loss function in Eqn.~(\ref{loss_func}).

To optimize the above Problem~(\ref{fed_obj}), we refer to the original FedAVG algorithm in \cite{mcmahan2017}. As shown in Fig.~\ref{s0}, the overall workflow of FedParking is described as follows.
\begin{itemize}
	\item Global Model Downloading: An application provider deploys a  parameter server to build and iteratively update a time series predict model as the global model for parking space estimation. When the global model is initialized, each PLO is notified to download it. 
	
	\item  Local Model Training: As required by the parameter server, each PLO performs a local model training task by the local data.  After that, they upload the updated model parameters to the parameter server.

	\item  Model Parameter Submission: All the model parameters are collected and aggregated by the parameter server to update the weight $\Theta_0$ and improve the accuracy of the global model.
	
	\item Global Model Updating: A new global  model is acquired and transmitted to all the PLOs. The iterations of local model training and aggregation repeats until the global model converges. 
\end{itemize}
Note that the whole training procedure of the global model uses the distributed training datasets of all the PLOs, and is executed through peer-to-peer collaboration without exchanging the raw data.

\subsection{LSTM for FedParking}
We apply a LSTM model as the important global model in FedParking. Compared with the original recurrent neural network, LSTM has advantages in remembering and forgetting the information in the time sequence. This enables us to achieve an accurate estimation of the parking spaces.

A LSTM network consists of two data gates to capture the sequence information, i.e., update gate and forget gate. In the following, we provide more details of the procedure of local model training. Take PLO $j$ as an example. By data collection, parking space information (e.g., parking occupancy rate) of a parking lot over different time slots is recorded. We apply a sliding window approach to extract the input time series ${X_j} = \{ {x_1},{x_2}, \cdots ,{x_{{{\left\| {{{\mathcal D}_j}} \right\|}}}}\}$  of the long short-term memory model from the data source. When all the original data samples consist of $y$ time slots, we  obtain  ${\left\| {{{\mathcal D}_j}} \right\|}=y-z+1$ input time series if the sliding window size is $z$. Let  $H_j=\{h_1, h_2, \cdots, h_{{{\left\| {{{\mathcal D}_j}} \right\|}}}\}$ indicate the hidden states, and $O_j=\{o_1, o_2, \cdots, o_{{{\left\| {{{\mathcal D}_j}} \right\|}}}\}$ represent the output states. We use $W^{\cdot\rm{h}}$ and $W^{\cdot\rm{x}}$ to denote the weights, and use $b^{\cdot}$ to denote the bias.

At the time step $t$, the value of forget gate $\Gamma _t^{\rm{f}}$ is calculated by
\begin{equation}
	\Gamma _t^{\rm{f}} = \sigma ({W^{\rm{fh}}}{h_{t - 1}} + {W^{\rm{fx}}}{x_t} + {b^{\rm{f}}}),
\end{equation}
where $x_t$ is the input vector at the time step $t$ and  $h_{t-1}$ is the hidden state at the previous time step, $\sigma (\cdot)$ denotes a Sigmoid function that normalizes the value into the range $\left[0, 1\right]$. The forget gate determines what information will be eliminated from the cell state. When the value of the forget gate $\Gamma _t^{\rm{f}}=0$, it throws away all the information, and $\Gamma _t^{\rm{f}}=1$ means that it keeps all the information.  As for the update gate, it decides what information can be stored in the cell state. The value of update gate is calculated by
\begin{equation}
	\Gamma _t^{\rm{u}} = \sigma ({W^{\rm{uh}}}{h_{t - 1}} + {W^{\rm{ux}}}{x_t} + {b^{\rm{u}}}).
\end{equation}

Next, we calculate the information of candidate memory cell ${\tilde c_t}$ at the time step $t$ by
\begin{equation}
	\tilde c_t = \tanh ({W^{\rm{ch}}}{h_{t - 1}} + {W^{\rm{cx}}}{x_t} + {b^{\rm{c}}}),
\end{equation}
where ${\tilde c_t}$ represents the candidate information that should be stored in the cell state. Until now, we obtain $\{\Gamma _t^{\rm{f}}, \Gamma _t^{\rm{u}}, \tilde c_t\}$ to update the cell state by
\begin{equation}
	{c_t} = \Gamma _t^{\rm{f}} \odot {c_{t - 1}} + \Gamma _t^{\rm{u}} \odot {\tilde c_t},
\end{equation}
where $\odot$ is an operator of executing the point-wise multiplication of two vectors. The output and hidden states are computed by
\begin{equation}
\left\{ \begin{array}{l}
{o_t} = \sigma ({W^{{\rm{oh}}}}{h_{t - 1}} + {W^{{\rm{ox}}}}{x_t} + {b^o})
\\
{h_t} = {o_t} \odot \tanh ({c_t}){\rm{.}}
\end{array} \right.
\end{equation}
Finally, we take $o_t$ as the input for the multilayer perceptron and compute the time series prediction $\hat{x}_{t+1}$ of the next time step. After the local model training, the weights $W^{\cdot\rm{h}}$ and $W^{\cdot\rm{x}}$, and the bias $b^{\cdot}$ are shared to the parameter server. 

With the help of FedParking, each PLO can predict the number of available parking space over time, which is  utilized as a parking capacity constraint to adjust the following incentive mechanism design when necessary. 
\section{Two-Stage Stackelberg Game Formulation for Incentive Mechanism Design}
Within a time period, a batch of parking requests are gathered and the PLOs are notified to process them.  In this paper, we model the interactions among the PLOs and vehicles as a  Stackelberg game, where each PLO plays as a game leader and provides monetary rewards for the vehicles in  Stage I, and each vehicle plays as a game follower and adopts a probabilistic strategy to decide the parking choice and the number of computing resources shared to a PLO in Stage II.
\subsection{Vehicles in Stage II}
Vehicle $i$ is incentivized to enter a parking lot $j$ and share its idle computing resources during the parking time. Let $r^j$ represent the crucial incentive parameter of PLO $j$ that indicates the monetary rewards  per unit computing resource and unit time.  According to the distance between the parking lot and destination, vehicle $i$ may have an original preference $p_i^j$ over parking lot $j$ and the reserved parking duration is $d_i$. We express the utility gain of vehicle $i$ by $p_i^jr^jf_i^jd_i$, where $f_i^j$ is the decision variable of vehicle $i$ referring to the number of computing resources shared to PLO $j$.  To support offloading services, vehicle $i$ is required by PLO $j$ to undertake workloads $w^j$. When processing the workloads, a majority of energy is consumed in the CPU execution. Similar to \cite{8648330}, we neglect the energy consumed in the data transmission. For vehicle $i$, the energy consumption cost of CPU execution is  ${{\kappa _i}} f_i^{j2}{w^j}$, where ${{\kappa _i}}$ denotes the energy coefficient of the vehicle and represents the effective switched capacitance relying on the chip architecture \cite{8234686, 9113721}. 
As a consequence, vehicle $i$ acquires a following utility function if accepting the employment of PLO $j$, 
\begin{equation}
	U_i^j=p_i^j{r^j}f_i^j{d_i} - {\kappa _i}f_i^{j2}{w^j}.
\end{equation} 

Motivated by the incentive mechanism design in the previous work \cite{9169846}, we consider that vehicle $i$ makes a probabilistic decision on the parking destination.  Particularly, we consider that monetary incentives have a relatively strong positive impact on the parking decision of the vehicle since the monetary rewards can compensate for the parking fee after it shares the idle computing resources to the PLO. We formulate the pairing probability between the vehicle and PLO by only the monetary rewards. Clearly, a vehicle is more likely to enter a parking lot with more monetary rewards. Thus, we set the pairing probability between vehicle $i$ and PLO $j$ as follows
\begin{equation}
	\rho_i^j={\frac{{{r^j}}}{{\sum\limits_{1 \le k \le J} {{r^k}} }}}.
\end{equation}
For vehicle $i$, we have $\sum\nolimits_{1 \le j \le J}{\rho_i^j}=1$. Considering $J$ PLOs, the expected utility of vehicle $i$ is
 \begin{equation}
 	{{\mathcal U}_i} ({{\bf{f}}_i}, {\bf{r}}) = \sum\limits_{1 \le j \le J} {\rho _i^jU_i^j},
 \end{equation}
 where ${{\bf{f}}_i} = \left\{ {f_i^1,\cdots,f_i^J} \right\}$ and ${\bf{r}} = \left\{ {{r^1},\cdots,{r^J}} \right\}$. Each vehicle $i$ decides the optimal strategy vector ${{\bf{f}}_i}$ to maximize ${\mathcal U}_i$ according to the reward policy vector ${\bf{r}}$.
 
\subsection{Parking Lot Operators in Stage I}
PLO $j$ collects a certain number of computing resources from the PVs and is able to provision offloading services. After satisfying the computation demand of offloading users, PLO $j$ obtains revenue $g^j$ per unit computing resource and unit time. Considering the total monetary cost for $I$ PVs, PLO $j$ obtains the expected utility in terms of profit as follows
\begin{equation}
\mathcal{V}^j({r^j}, {{\bf{r}}^{ - j}}, {{\bf{f}}^j}) = \sum\limits_{1 \le i \le I} {\rho _i^j({g^j} - {r^j})f_i^j{d_i}}, 
\end{equation}
where ${{\bf{r}}^{ - j}}$ is  a reward policy vector released by all the PLOs except PLO $j$ and ${\bf{f}}^j=\left\{f_1^j,\cdots,f_I^j\right\}$. Each PLO $j$ decides the optimal strategy $r^j$ to maximize ${\mathcal V}^j$ according to the strategies of all the other PLOs (i.e., ${\bf{r}}^{-j}$) and the strategies of all the vehicles (i,e., ${\bf{f}}^j$). Note that there exists a non-cooperative game among the $J$ PLOs in the incentive mechanism. Moreover, two essential constraints of $r^j$ are presented. The first constraint named by parking capacity constraint is expressed by  $\sum\nolimits_{1 \le i \le I} {\rho _i^j}  \le {n^j}$, where $n^j$ is acquired from the above FedParking scheme and represents the predicting number of available parking spaces in the parking lot. This means that  PLO $j$ should adjust $r^j$ well to control the expected number of arriving vehicles with respect to the given parking capacity constraint. Considering a budget constraint, $r^j$ is constrained within an upper limit $r_{\max }^j \le g^j$.

Each player (i.e., each PLO and vehicle) of the Stackelberg game has its own interest. We aim to find a unique Stackelberg equilibrium at which each leader (PLO)  obtains the maximal expected utility given the best responses from the followers (vehicles). At the time, no single PLO or vehicle will have any motivation to unilaterally change its decision. Given the multiple-leader multi-follower Stackelberg game, we define the Stackelberg equilibrium  as follows.

\textbf{Definition 1 (Stackelberg Equilibrium).} There exist an optimal computing resource vector denoted as ${{\bf{f}}_i^*}$ and an optimal monetary reward $r^{j*}$ for vehicle $i$ and PLO $j$, respectively. $({{\bf{f}}^*} = \left\{ {{\bf{f}}_i^*} \right\},{{\bf{r}}^*} = \left\{ {{r^{1*}},\cdots,{r^{J*}}} \right\}, \forall i$ and $j)$  is the Stackelberg equilibrium if and only if the following conditions are satisfied.
\begin{equation*}
\left\{ \begin{array}{l}
	{{\mathcal V}^j}({r^{j*}},{{\bf{r}}^{ - j*}},{{\bf{f}}^{j*}}) \ge {{\cal V}^j}({r^j},{{\bf{r}}^{ - j*}},),\forall j,\\
	{{\mathcal U}_i}({\bf{f}}_i^*,{{\bf{r}}^*}) \ge {{\cal U}_i}({{\bf{f}}_i},{{\bf{r}}^*}),\forall i.
\end{array} \right.
\end{equation*}

We find the subgame perfect Nash equilibrium to achieve the Stackelberg equilibrium, which ensures that the expected utility of each PLO is maximized subject to  the best responses from the vehicles. In the proposed Stackelberg game, there exists a subgame among the PLOs in the upper stage and they strictly compete in a non-cooperative fashion. The Nash equilibrium of the subgame is denoted as $(r^{j*},{{\bf{r}}^{- j*}})$, where the expected utility of each PLO $j$ cannot be further improved by adopting a different strategy other than the given strategies of the other PLOs ${\bf{r}}^{ - j*}$. In the bottom stage, the best response of vehicle $i$ is acquired as $\bf{f}_i^*$ by maximizing the expected utility with respect to $\bf{r}^*$.  To reach the Stackelberg equilibrium, we adopt the backward induction method in the following.

\section{Stackelberg Equilibrium Analysis Under Complete Information}
We temporarily neglect the parking capacity constraints and provide the theoretical analysis to prove the existence and uniqueness of the above Stackelberg equilibrium. 
\subsection{Best Response of a Vehicle}
We first pay attention to the best response of a follower. Given the reward policy vector $\bf{r}$ published by all the PLOs, each vehicle $i$ determines $\bf{f}_i$ to maximize its $\mathcal{U}_i$. We take the first-order and second-order derivatives of $\mathcal{U}_i$ with respect to $f_i^j$,
\begin{equation*}
	\begin{array}{l}
		\frac{{\partial {{\mathcal U}_i}}}{{\partial f_i^j}} = \rho _i^j(p_i^j{r^j}{d_i} - 2{\kappa _i}f_i^j{w^j}),\\
		\frac{{{\partial ^2}{{\mathcal U}_i}}}{{\partial f_i^{j2}}} =  - 2\rho _i^j{\kappa _i}{w^j} < 0.
	\end{array}
\end{equation*}
The negative second-order derivative of ${\mathcal U}_i$ indicates that ${\cal U}_i$ is strictly concave with respect to $f_i^j$.  As a consequence, for vehicle $i$, the best response $f_i^{j*}$ can be calculated by
\begin{equation}
	f_i^{j*} = \frac{{p_i^j}{d_i}}{{2{\kappa _i}{w^j}}}{r^j} = \lambda _i^j{r^j}, \label{optf} 
\end{equation}
where $\lambda_i^j=p_i^jd_i/(2{\kappa _i}{w^j})$. The above result means that  more monetary rewards are provided to the vehicle, more computing resources will be shared to the PLO, which is consistent with the intuition.
\subsection{Nash Equilibrium Analysis for PLOs}
We realize that each PLO is selfish to employ the vehicles with idle computing resources and maximize the payoffs. The PLOs play a non-cooperative game because of the competition effect. In the non-cooperative game, the player set consisting of  $J$ PLOs is finite and strategy spaces of the PLOs are nonempty, convex, and compact subsets of the Euclidean spaces. The Nash equilibrium exists in the non-cooperative game if we can prove the concavity of the utility functions in the strategy spaces.

\textit{Theorem 1}. Under the assumption that ${3\sum\nolimits_{k \ne j} {{r^k}} > {g^j}}, \forall j$,  a Nash equilibrium exists in the non-cooperative game among the PLOs.  

\textit{Proof:} By substituting $f_i^{j*}$ into $\mathcal{V}^j$, we update
\begin{equation}
	\begin{aligned}
			{{\mathcal V}^j}& = \frac{{{g^j}{r^{j2}}\sum\limits_{1 \le i \le I} {\lambda _i^j{d_i}}  - {r^{j3}}\sum\limits_{1 \le i \le I} {\lambda _i^j{d_i}} }}{{\sum\limits_{1 \le k \le J} {{r^k}} }} \\
		&= {\mu ^j}\frac{{{g^j}{r^{j2}} - {r^{j3}}}}{{\sum\limits_{1 \le k \le J} {{r^k}} }},
	\end{aligned}
\end{equation}
where ${\mu ^j} = \sum\nolimits_{1 \le i \le I} {\lambda _i^j{d_i}} $. We take the first-order and second-order derivatives of $\mathcal{V}^j$ with respect to $r^j$, 
\begin{equation*}
		\begin{array}{*{20}{l}}
			\begin{array}{l}
				\frac{{\partial {V^j}}}{{\partial {r^j}}} = {\mu ^j}\frac{{(2{g^j}{r^j} - 3{r^{j2}})\sum\limits_{1 \le k \le J} {{r^k}}  - {g^j}{r^{j2}} + {r^{j3}}}}{{{{\left( {\sum\limits_{1 \le k \le J} {{r^k}} } \right)}^2}}}\\
				~~~~ = {\mu ^j}\frac{{{r^j}\left[ { - 2{r^{j2}} - \left( {3\sum\limits_{k \ne j} {{r^k}}  - {g^j}} \right){r^j} + 2{g^j}\sum\limits_{k \ne j} {{r^k}} } \right]}}{{{{\left( {\sum\limits_{1 \le k \le J} {{r^k}} } \right)}^2}}} \\
				~~~~= {\mu ^j}\frac{{{r^j}{\Pi ^j}}}{{{{\left( {\sum\limits_{1 \le k \le J} {{r^k}} } \right)}^2}}},
			\end{array}\\
			{\frac{{{\partial ^2}{V^j}}}{{\partial {r^{j2}}}} =  - 2{\mu ^j}\frac{{{r^j}{\Pi ^j}}}{{{{\left( {\sum\limits_{1 \le k \le J} {{r^k}} } \right)}^3}}} + {\mu ^j}\frac{{{\Pi ^j} - 4{r^j} - \left( {3\sum\limits_{k \ne j} {{r^k}}  - {g^j}} \right)}}{{{{\left( {\sum\limits_{1 \le k \le J} {{r^k}} } \right)}^2}}}.}
		\end{array}
\end{equation*}
where ${{\Pi ^j}}={ - 2{r^{j2}} - (3\sum\nolimits_{k \ne j} {{r^k}}  - {g^j}){r^j} + 2{g^j}\sum\nolimits_{k \ne j} {{r^k}} }$.   If $\partial{\mathcal{V}^j}/\partial{r^j}=0$, $\Pi^j =0$, at this time, ${\partial ^2}{\mathcal{V}^j}/{\partial ^2}{r^{j2}}<0$ since  ${3\sum\nolimits_{k \ne j} {{r^k}} > {g^j}}$ and  $r^j >0$. This means that utility function of PLO $j$ is quasi-concave. We conclude that the Nash equilibrium exists in this non-cooperative game. $\blacksquare$

\textit{Theorem 2}. The Nash equilibrium of the non-cooperative game among the PLOs is unique.

\textit{Proof:}  Based on the first-order optimality condition $\partial {\mathcal{V}^j}/\partial {r^j}=0$ (i.e., ${\Pi ^j}=0$), we have ${r^{j*}}={\phi ^j}({\bf{r}})$, which is shown at the top of the next page.
Considering the upper limit of $r^j$, we update 
\begin{equation}
	{r^{j*}} = \left\{ \begin{array}{l}
		{\phi ^j}({\bf{r}}),~0 < {\phi ^j}({\bf{r}}) < r_{\max }^j,\\
		r_{\max }^j,~{\phi ^j}({\bf{r}}) \ge r_{\max }^j.
	\end{array} \right.
\end{equation}
Next, we pay attention to the case that $r^{j*} < r^{j}_{\max}$. We prove the uniqueness of the Nash equilibrium by validating that the best response function $\phi^j({\bf{r}})$ is a standard function, which satisfies three properties as follows
\begin{itemize}
	\item \textit{Positivity:} $\phi^j({\bf{r}}) >0$.
	\item \textit{Monotonicity:} if $\bf{r}^{\prime}>\bf{r}$, then $\phi^j({\bf{r}^{\prime}}) > \phi^j({\bf{r}})$.
	\item \textit{Scalability:} for any $\alpha>1$, $\alpha\phi^j({\bf{r}}) > \phi^j({\alpha\bf{r}})$.
\end{itemize}
First, due to $g^j>0$ and $\bf{r}> \bf{0}$, $\phi^j({\bf{r}}) >0$ is always satisfied. Second, 
we easily know that 
\begin{equation*}
	{\left(3\sum\limits_{k \ne j} {{r^k}}  + \frac{{5}}{3}{g^j}\right)^2} >  {{{\left(3\sum\limits_{k \ne j} {{r^k}}  - {g^j}\right)}^2} + 16{g^j}\sum\limits_{k \ne j} {{r^k}} }.
\end{equation*}
Furthermore, 
\begin{equation*}
	\frac{{\partial {\phi ^j}({\bf{r}})}}{{\partial \left( {\sum\limits_{k \ne j} {{r^k}} } \right)}} = \frac{{0.25\left( {18\sum\limits_{k \ne j} {{r^k}}  + 10{g^j}} \right)}}{{2\sqrt {{{\left( {3\sum\limits_{k \ne j} {{r^k}}  - {g^j}} \right)}^2} + 16{g^j}\sum\limits_{k \ne j} {{r^k}} } }} - 3 > 0.
\end{equation*}
When  $\bf{r}^{\prime}>\bf{r}$,  $\sum\nolimits_{k \ne j} {{r^{k\prime}}}  > \sum\nolimits_{k \ne j} {{r^k}}$ and we have $\phi^j({\bf{r}^{\prime}}) > \phi^j({\bf{r}})$. Finally, $\forall \alpha >1$, we have $\alpha^2 > \alpha$ and further have $\alpha {\phi ^j}({\bf{r}}) > {\phi ^j}(\alpha {\bf{r}})$, which is shown at the top of the next page.
%

\begin{figure*}[!t]	
	\normalsize	
	\setcounter{mytempeqncnt}{\value{equation}}	
	\setcounter{equation}{13}
	\begin{equation*}
		{r^{j*}} = 0.25\left( {\sqrt {{{\left(3\sum\limits_{k \ne j} {{r^k}}  - {g^j}\right)}^2} + 16{g^j}\sum\limits_{k \ne j} {{r^k}} }  - 3\sum\limits_{k \ne j} {{r^k}}  + {g^j}} \right)= {\phi ^j}({\bf{r}}).
	\end{equation*}	
	\begin{equation*}
		\begin{aligned}
\alpha {\phi ^j}({\bf{r}})&= 0.25\sqrt {9{\alpha ^2}{{\left( {\sum\limits_{k \ne j} {{r^k}} } \right)}^2} + {\alpha ^2}{g^{j2}} + 10{\alpha ^2}{g^j}\sum\limits_{k \ne j} {{r^k}} }  - 3\alpha \sum\limits_{k \ne j} {{r^k}}  + {g^j} \\
&> 0.25\sqrt {9{\alpha ^2}{{\left( {\sum\limits_{k \ne j} {{r^k}} } \right)}^2} + {g^{j2}} + 10\alpha {g^j}\sum\limits_{k \ne j} {{r^k}} }  - 3\alpha \sum\limits_{k \ne j} {{r^k}}  + {g^j} = {\phi ^j}(\alpha {\bf{r}}).		
\end{aligned}
	\end{equation*}		
	\setcounter{equation}{\value{mytempeqncnt}}	
	\hrulefill	
	\vspace*{4pt}
\end{figure*}

Note that we have proved that the best response function of PLO $j$ satisfies the above three properties in this case. When $r^{j*} =r^{j}_{\max}$, these three properties are still satisfied. Then we conclude that standard best response functions of all the PLOs have the unique Nash equilibrium.
$\blacksquare$

\textit{Theorem 3}. There exists a unique Stackelberg equilibrium in the Stackelberg game.

\textit{Proof:} Until now, we have proved that the best response of a follower is unique in the second stage and the Nash equilibrium among the leaders is also unique in the first stage. We can conclude that the Stackelberg equilibrium is unique. The proof is completed. $\blacksquare$

To reach the Stackelberg equilibrium,  we utilize the following rule to update the monetary rewards in each iteration $k$ to gradually approach the Nash equilibrium in the first stage,
\begin{equation}
	{r^{j, k}} = {r^{j,k - 1}} + {\lambda ^j}{r^{j, k-1}}\frac{{\partial {\mathcal{V}^j}}}{{\partial {r^j}}},~\forall j,
\end{equation}
where $\lambda^j$ denotes a learning rate,  and  $\partial {\mathcal{V}^j}/{{\partial {r^j}}}$ is approximately calculated by the central difference method,
\begin{equation}
\frac{{\partial {{\cal V}^j}}}{{\partial {r^j}}} \approx \frac{{{\mathcal V}_ + ^{j,k} - {\mathcal V}_ - ^{j,k}}}{{2\Delta r}},
\end{equation}
where ${\mathcal V}_ + ^{j,k} = {{\mathcal V}^j}({r^{j,k - 1}} + \Delta r,{{\bf{r}}^{ - j,k - 1}})$, ${\mathcal V}_ - ^{j,k}={{\mathcal V}^j}({r^{j,k - 1}} - \Delta r,{{\bf{r}}^{ - j,k - 1}})$ and $\Delta r $ is a small constant indicating the minor change of the reward policy. Given the updating rule of the monetary rewards, two different approaches can be considered by the PLOs to iteratively adjust their strategies, including Gauss-Seidel and Jacobi based algorithms. In the former algorithm, the PLOs update their strategies sequentially. In the latter algorithm, they update the strategies in parallel. In this paper, we adopt the Jacobi based algorithm.

\section{A Learning based Incentive Mechanism by DRL}
In Section V, we focus on analyzing the Stackelberg equilibrium under complete information, without taking into account the parking capacity constraints. However, both the arrivals of the vehicles and number of available parking spaces in each parking lot are varying over time. It is difficult to obtain the closed-form Stackelberg equilibrium in practical scenarios. In addition, the above theoretical analysis is based on the centralized decision making which requires to collect private information of all the players to derive the solution of the Stackelberg game. This violates the individual privacy of the players.  Under the complicated decision-marking conditions and for the fulfillment of privacy-related requirements,  advanced artificial intelligence technologies can be potential tools to facilitate the incentive mechanism design \cite{8926369,8944281}.  Hence, we are motivated by \cite{8963610} to apply DRL as an online algorithm to tackle the incentive mechanism design problem of the multi-leader multi-follower Stackelberg game under incomplete information.  The extensively concerned DRL technique has outstanding ability in  adapting to the non-static environment and dynamically resolving the complicated decision-making problems according to environment variations \cite{8892573}. Our DRL approach enables each PLO as an agent to quickly learn a near-optimal decision in a distributed and privacy-friendly manner.  We first outline a learning framework of the whole DRL based incentive mechanism. After that, we formulate the incentive design problem as a DRL learning task and provide more details of the neural network configuration and optimization method based on the proposed learning framework.

\begin{figure}
\centering
 \includegraphics[width=0.48\textwidth]{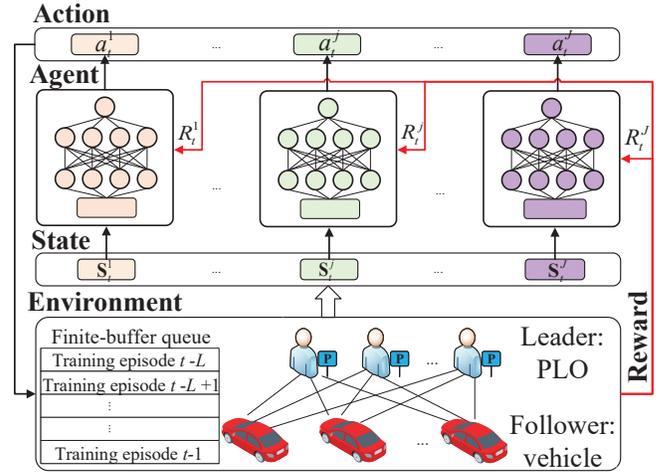}
 \caption{A learning based incentive mechanism by multi-agent DRL.}
 \label{workflow}
\end{figure}

\subsection{Overview of the Learning Framework}
To reach the Stackelberg equilibrium, each PLO becomes an agent in the DRL approach reacting to the environment consisting of all the vehicles, as shown in Fig.~\ref{workflow}.  PLO $j$ is  agent $j$, which refers to the historical strategies of the other agents and determines an action according to the current state. 

Specifically, PLO $j$ in each training episode $t (1\le t \le T)$ interacts with the environment to know the current state ${\bf{S}}^j_t$, which consists of reward policy history of the other PLOs and its profit parameter. PLO $j$ plays as a game leader that decides the action $a_t^j$ and releases the reward policy $p_t^j$. After that, a vehicle with parking demand in the environment plays as a follower that confirms the optimal strategy based on Eqn.~(\ref{optf}). By collecting the strategies of all the vehicles, the environment computes the learning rewards $R_t^j$ for each PLO. At the same time, the environment saves all the historical data of the actions into a finite-buffer queue.  Let $L$ denote the size of the finite-buffer queue. Here,  an application provider of the parking services in the region could become such an essential coordinator to establish the learning environment. The application provider is convenient to collect historical strategies of each player and accordingly update the finite-buffer queue. It  extracts the related information from the finite-buffer queue to generate new states for the corresponding PLOs. Hence, the next training episode is conducted. 

Note that during the training procedure, each player cannot obtain any privacy-related information of the other players that influences their decisions. As a consequence, we solve the Stackelberg game without leaking private information and causing privacy threats to the players.
\subsection{Design Details}
In the following, we describe a DRL learning task under the proposed learning framework. For each player, the decision-making procedure is performed independently according to the others’ past strategy information and its own property. We further formulate the interactions among the PLOs as a multi-agent markov decision process.

\paragraph{State space}
We focus on a case where state space of each player is partially observable. For PLO $j$, the state is formed by the past strategies of the other PLOs and its own properties, which is denoted as $\{\mathbf{r}^{-j}_{t-L}, \mathbf{r}^{-j}_{t-L+1},\cdots,\mathbf{r}^{-j}_{t-1}\}$ and $g^j$, respectively. In the training episode $t$,  the PLO interacts with the environment to know the current state, which is represented by $\mathbf{S}_t^j=\{\mathbf{r}^{-j}_{t-L}, \mathbf{r}^{-j}_{t-L+1}, \cdots, \mathbf{r}^{-j}_{t-1}, g^j\}$. 

\paragraph{Action space}
We use $\mathcal{A}=\{a^j_t, \forall j\}$ to represent the actions of all PLOs. In the training episode $t$, PLO $j$ determines the action $a_t^j$ with respect to the state $\mathbf{S}_t^j$. The action refers to the reward policy. To improve the learning efficiency, we set lower and upper bounds $\left[r^j_{\min}, r^j_{\max}\right] $ for the action. Since $a_t^j\in [r^j_{\min}, r^j_{\max}]$,  the action space of $a_t^j$  is continuous and we simply normalize the range of $a_t^j$ to $[0, 1]$ by letting $a^j_t=(r^j_t - r^j_{\min})/ r^j_{\max}$.

\paragraph{Reward function}
The reward space is defined as $\mathcal{R}=\{R^j_t, \forall j\}$. We consider a time-varying parking capacity constraint for each agent. In the training episode $t$, $n_t^j$ is the available parking capacity of agent $j$. This means that for all the $I$ vehicles, the expected number of the vehicles choosing PLO $j$ is limited by $n_t^j$. With respect to the parking capacity constraint, we are motivated to express the reward function of agent $j$ as 
\begin{equation}
R_t^j = \left\{ \begin{array}{l}
\sum\limits_{1 \le i \le I} {\rho _i^j({g^j} - r_t^j)f_i^j{d_i}},~\sum\limits_{1 \le i \le I} {\rho _i^j}  \le n_t^j\\
\frac{{n_t^j}}{I}\sum\limits_{1 \le i \le I} {({g^j} - r_t^j)f_i^j{d_i}}-{\mathcal{P}(\sum\limits_{1 \le i \le I} {\rho _i^j} )}, ~{\rm{otherwise}}{\rm{.}}
\end{array} \right.
\end{equation}
where $\mathcal{P}(\cdot)$ is a penalty function to control the expected number of arriving vehicles. The function is predefined by
\begin{equation}
	\mathcal{P} = \frac{{{\iota ^j}}}{{n_t^j}}\max (\sum\limits_{1 \le i \le I} {\rho _i^j}  - n_t^j,0),
\end{equation}
where $ {\iota ^j}$ is a penalty factor.
\paragraph{Learning Objective}
We first denote the reward policy of PLO $j$ parameterized by $\theta ^j$ as $\pi _{\theta}^j$, which is defined as $\pi _{\theta}^j: \mathbf{S}^j \rightarrow a^j$. Let $V_{\pi _{\theta}}^j(\mathbf{S}^j)$ indicate  the value function for the given state, and $Q_{\pi _{\theta }}^j(\mathbf{S}^j, a^j)$ become the value function for the given state and action. The learning objective of agent $j$ is shown by
\begin{equation}\label{learn_obj}
\begin{aligned}
\theta _*^j &= \mathop {\arg \max }\limits_{{\theta ^j}} {L_j}(\pi _\theta ^j)\\
 &= \mathop {\arg \max }\limits_{{\theta ^j}} {\mathbb E}\left[ {V_{{\pi _\theta }}^j(\mathbf{S}^j_0)} \right]\\
 &= \mathop {\arg \max }\limits_{{\theta ^j}} {\mathbb E}\left[ {Q_{{\pi _\theta }}^j(\mathbf{S}^j_0, a^j_0)\left| \pi _{\theta}^j \right.} \right],
\end{aligned}
\end{equation}
where
\begin{equation}
\begin{array}{l}
V_{{\pi _\theta }}^j(\mathbf{S}^j) = {\mathbb E}\left[\overline{R}_t^j \left| \mathbf{S}^j_t=\mathbf{S}^j, \Pi\right.\right],\\
Q_{{\pi _\theta }}^j(\mathbf{S}^j, a^j) = {\mathbb E}\left[\overline{R}_t^j \left| \mathbf{S}^j_t=\mathbf{S}^j, a^j_t=a^j, \Pi\right.\right],\\
\overline{R}_t^j=\sum\limits_{k = t}^T {{\gamma ^{k - t}}R_t^j}.
\end{array}
\end{equation}
Here,  $\Pi=\{ \pi _{\theta}^j, \forall j\}$ represents the whole policy set of all PLOs,  $\overline{R}_t^j$ is the expected discounted reward for PLO $j$ in all the $T$ training episodes, and $\gamma \in [0, 1]$ indicates the discounted factor.

\begin{figure*}[!htbp]
	\setlength{\belowcaptionskip}{-0.3 cm}
	\centering
	\subfigure[Parking lot operator 1.]{
		\label{s1a} 
		\begin{minipage}[!htbp]{0.3\linewidth}
			\centering
			\includegraphics[width=1.0\textwidth]{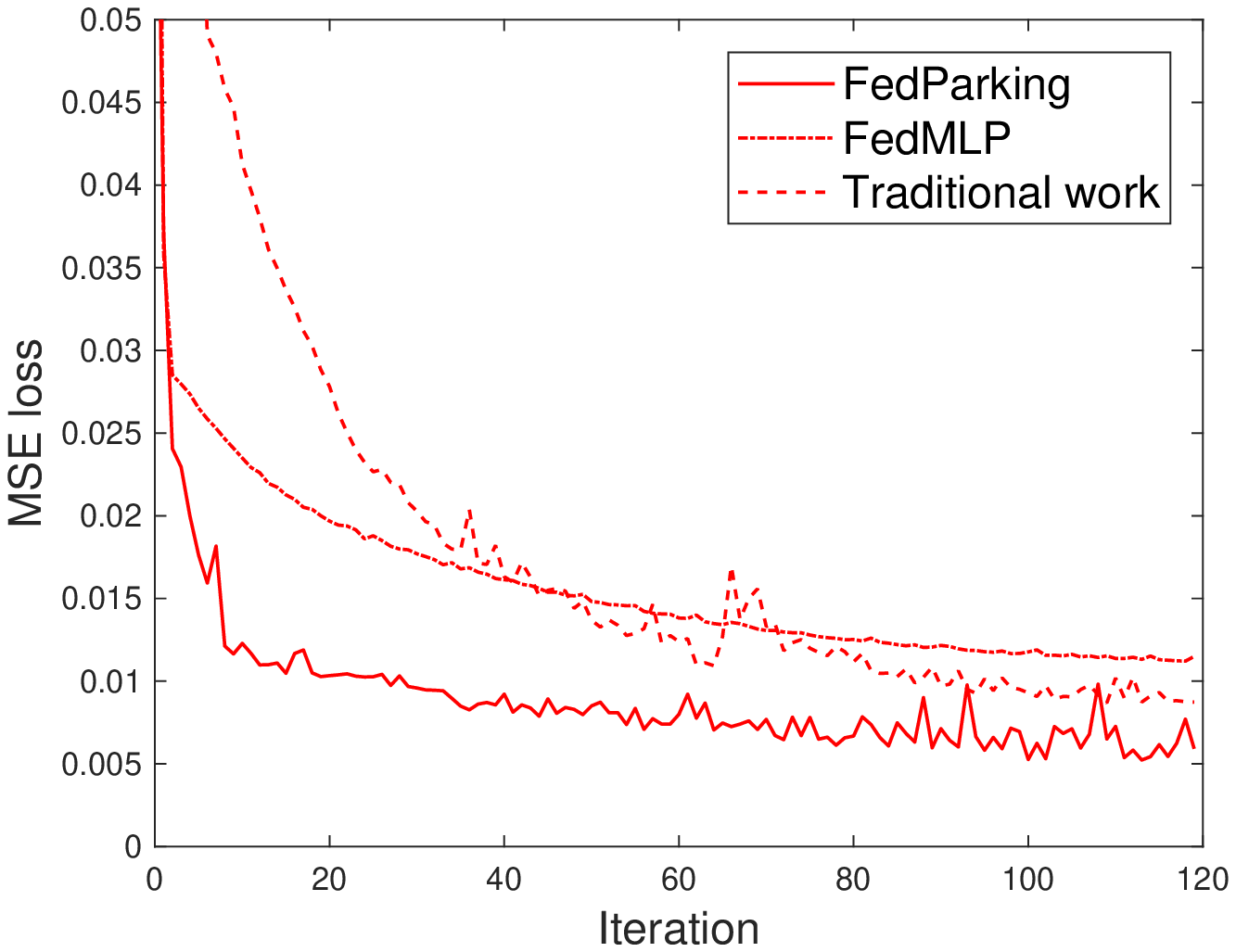}
	\end{minipage}}
	\subfigure[Parking lot operator 2.]{
		\label{s1b} 
		\begin{minipage}[!htbp]{0.3\linewidth}
			\centering
			\includegraphics[width=1.0\textwidth]{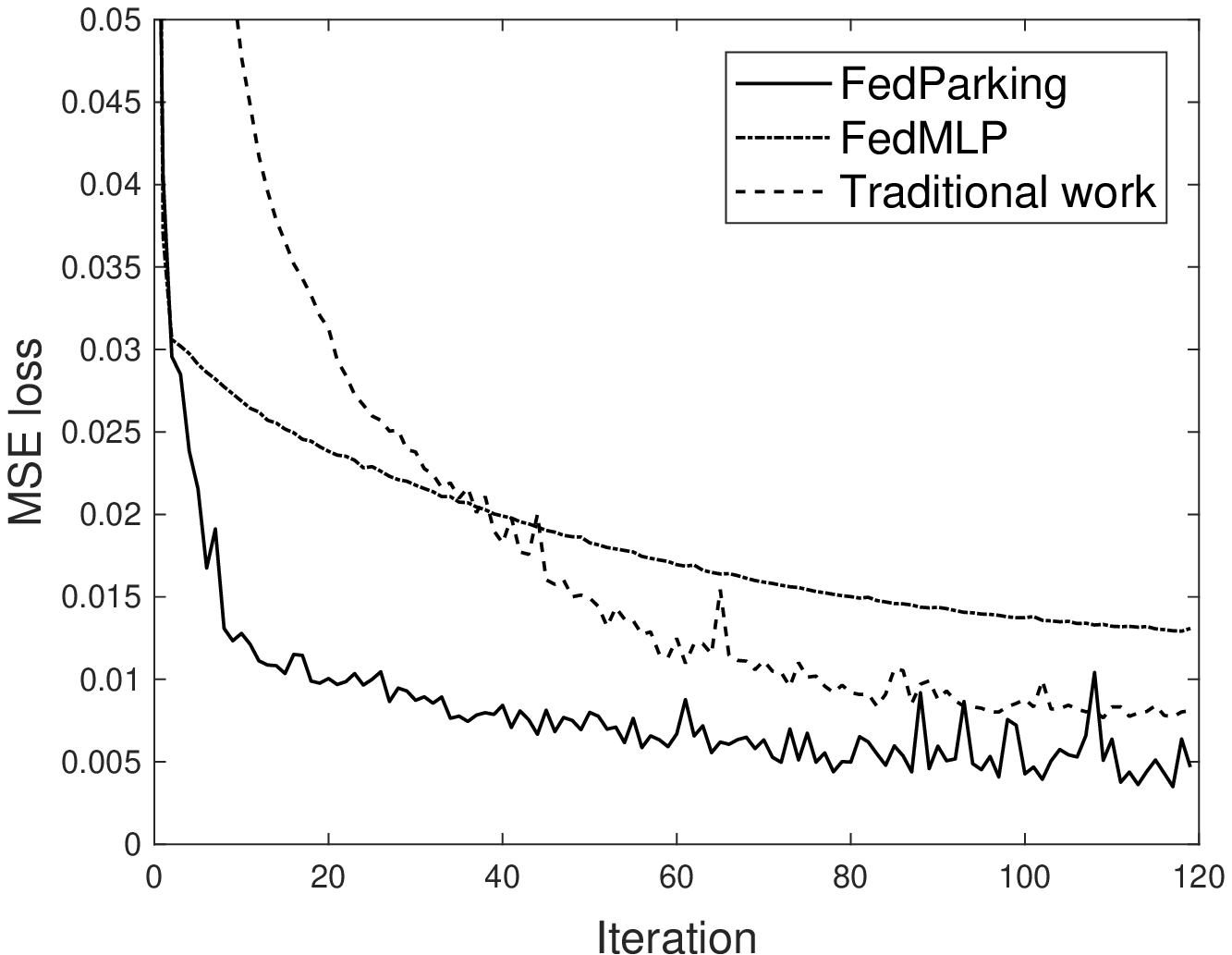}
	\end{minipage}}
	\subfigure[Parking lot operator 3.]{
		\begin{minipage}[!htbp]{0.3\linewidth}
			\centering
			\label{s1c} 
			\includegraphics[width=1.0\textwidth]{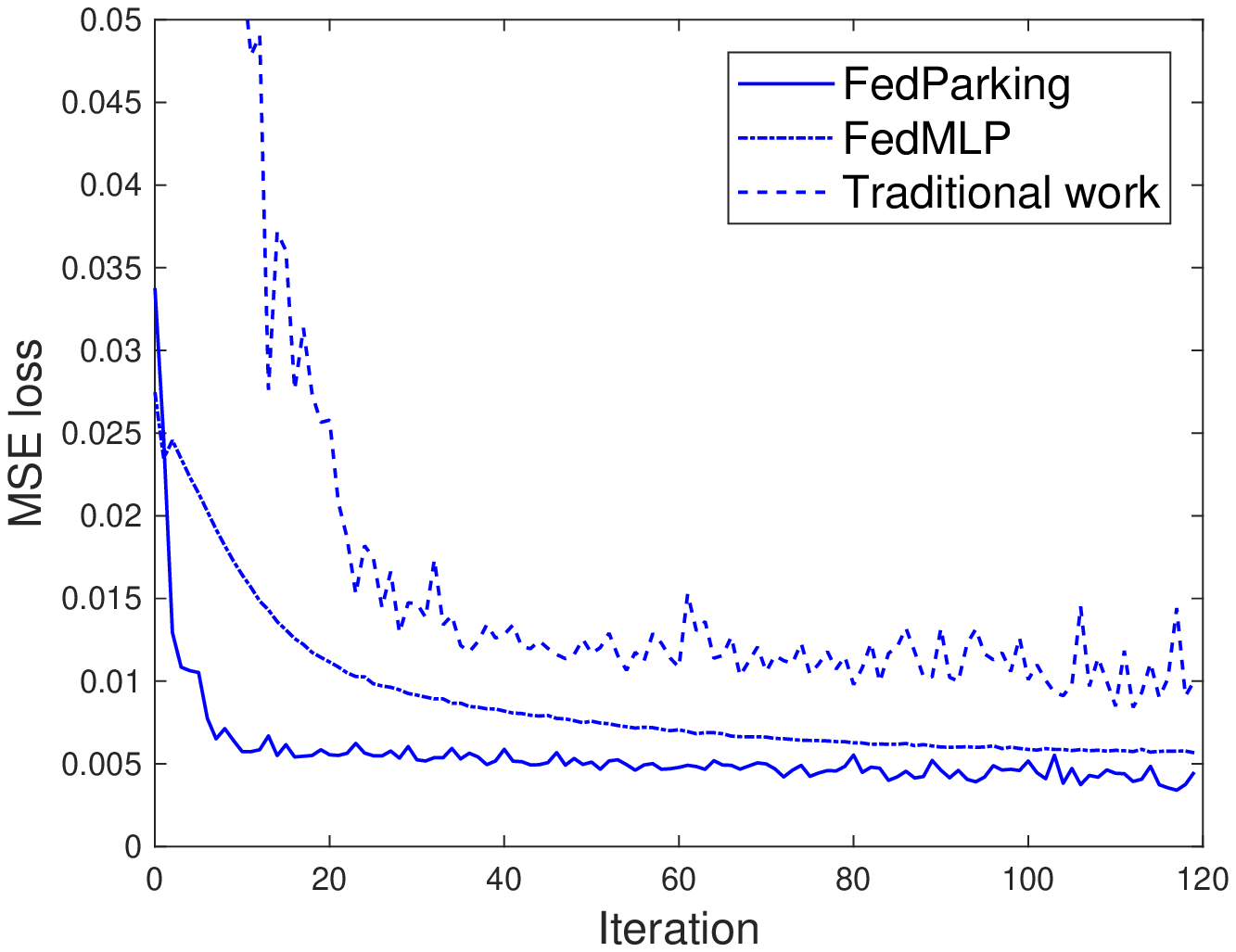}
	\end{minipage}}
	\caption{Comparison of training loss under different approaches.}
	\label{s1}
\end{figure*}

\subsection{Actor-Critic Networks and Policy Optimization}
We train the above  DRL model based on the policy-based methods and resorts to the typical actor-critic model.  We design a decentralized actor network and a decentralized critic network for each agent. For agent $j$, it maintains actor and critic networks parameterized by  $\theta ^j$ and $\omega ^j$ for approximating the policy and value function, respectively. We apply the policy gradient method to optimize the learning objective defined in Eqn.~(\ref{learn_obj}). Referring to the policy optimization theorem proposed in \cite{sutton1999policy}, the policy gradient is computed by
\begin{equation}\label{learn_grad}
\begin{aligned}
{\nabla _{{\theta ^j}}}{L^j} &= {E_{\pi _\theta ^j({\mathbf{S}^j})}}[{\nabla _{{\theta ^j}}}\log \pi _\theta ^j({\mathbf{S}^j},{a^j})A_{{\pi _\theta }}^j({\mathbf{S}^j},{a^j})]\\
&\approx {E_{\pi _\theta ^j({\mathbf{S}^j})}}[P^j{\nabla _{{\theta ^j}}}\log \pi _\theta ^j({\mathbf{S}^j},{a^j})A_{{\pi _\theta }}^j({\mathbf{S}^j},{a^j})],
\end{aligned}
\end{equation}
where $P^j=\frac{{\pi _\theta ^j({{\mathbf{S}^j}}\left|{a^j}\right.)}}{{ {\hat\pi _\theta ^j}({{\mathbf{S}^j}}\left|{a^j}\right.)}}, A_{{\pi _\theta }}^j({\mathbf{S}^j},{a^j})=Q_{{\pi _\theta }}^j(\mathbf{S}^j, a^j)-V_{{\pi _\theta }}^j(\mathbf{S}^j)$. $A_{{\pi _\theta }}^j({\mathbf{S}^j},{a^j})$ is the advantage function for state and action and ${\hat\pi _\theta ^j}({{\mathbf{S}^j}},{a^j})$ is the policy for importance sampling. Furthermore, we design the training procedure according to the proximal policy optimization algorithm in \cite{schulman2017}, which is the state-of-the-art policy gradient method with excellent stability. Therefore, the policy gradient is further clipped as 
\begin{equation}\label{learn_grad_clip}
{\nabla _{{\theta ^j}}}{L^j} \approx {E_{\pi _\theta ^j({\mathbf{S}^j})}}[{\nabla _{{\theta ^j}}}\log \pi _\theta ^j({\mathbf{S}^j},{a^j})\mathcal{C}_{{\pi _\theta }}^j({\mathbf{S}^j},{a^j})]
\end{equation}
where 
\begin{equation}
\mathcal{C}_{{\pi _\theta }}^j({\mathbf{S}^j},{a^j}) = \min [P^j A_{{\pi _\theta }}^j({\mathbf{S}^j},{a^j}), \mathcal{F} (P^j)A_{{\pi _\theta }}^j({\mathbf{S}^j},{a^j})],
\end{equation}
\begin{equation}
\mathcal{F} (P^j) = \left\{ {\begin{array}{*{20}{c}}
{1 + \varepsilon ,}&{P^j > 1 + \varepsilon }\\
{P^j,}&{1 - \varepsilon  \le P^j \le 1 + \varepsilon }\\
{1 - \varepsilon ,}&{P^j < 1 - \varepsilon }
\end{array}} \right.
\end{equation}
and $\varepsilon$ is an adjustable hyper-parameter. After that, we update the actor and critic models by the mini-batch stochastic gradient ascent and descent methods, respectively.  

In Fig.~\ref{workflow}, the actor network of a PLO is the multilayer perceptron with two hidden layers, consisting of 200 neurons and 50 neurons, respectively. For the critic network, we combine connected-layer and differentiable neural computer model in \cite{graves2016} to build the whole model for value estimation. Differentiable neural computer model is introduced as a type of recurrent neural network with internal memory module, and has been successful applied by the previous works to solve the partially observable markov decision process problem. The agents gradually learn the optimal strategies with the training procedure continuing. When the training procedure converges, they determine the strategies based on the outputs of the actor networks.

\section{Numerical Results}
\subsection{Parameter Setting}
We evaluate the proposed FedParking by numerical studies. First, we study the overall performance of FedParking on the real-world dataset named by Birmingham  parking  dataset \footnote{https://archive.ics.uci.edu/ml/datasets/Parking+Birmingham}. The dataset includes the parking history operated by National Car Parks in Birmingham, United Kingdom, and is updated every 30 minutes from 8:00 AM to 5:00 PM. In our experiments, we select 3 parking lots, referring to  ``BHMEURBRD01", ``BHMEURBRD02" and ``Bull Ring", to collaboratively learn a shared LSTM model for parking space estimation. A local dataset of a PLO is partitioned
using a 80/20 training/testing split. For a LSTM model as the global model in FedParking, the size of hidden state is 256. To extract the input time series of the LSTM model, we set the sliding window size $z=15$. 

A PLO acts as an offloading service provider and employs PVs as edge computing nodes. When a PV is effectively stimulated to share its idle computing resources, the PV continuously receives offloading tasks within a specific time period whose duration ranges from 5 to 20 minutes. The arrival process of the tasks is a typical Poisson process, where the average arrival rate and workload of the tasks are chosen from $\rm{U}\left[1, 3\right]$ per minute and $\rm{U}\left[2, 5\right]$ giga CPU cycles, respectively. We assume that there are 35 vehicles with parking demand. For a PV, reserved parking time $d$, computing capability $f$ and hardware parameter $\kappa$  are uniformly distributed in the range of ${\rm{U}}\left[20, 100\right]$ minutes, ${\rm{U}} \left[0.5, 3.5\right]$ GHz and  ${\rm{U}}\left[1, 10\right]$e-28, respectively. The preference parameter $p$ over different parking lots is randomly generated within $(0,1)$. For a PLO, profit parameter $g$ follows the uniform distribution $\rm{U}\left[3,5\right]$. As for the DRL experiments, we conduct them using Tensorflow 1.14 on Ubuntu 20.04 LTS with CUDA 10.1. We set the parameters $L=5$, $r_{\min}=0.2$, $r_{\max}=3$, $\iota=2$, $T=20$, $\gamma=0.95$ and $\varepsilon=0.1$ by default.

\subsection{FedParking for Parking Space Estimation}
In this paper, we study an integration of federated learning and LSTM to design FedParking, in which all PLOs collaborate to train a shared LSTM model for parking space estimation, ultimately improving the prediction accuracy while avoiding data transfer among the PLOs. In addition to FedParking, a general approach based on federated learning for parking space estimation is named by FedMLP, which adopts a multilayer perceptron model as the global model in federated learning. In the traditional work, each PLO only utilizes the local data for isolated training of its LSTM model. To evaluate the performance of our proposed approach in comparison with other approaches, we set the local batch size as 64. The local epoch is set as 1 for FedParking and FedMLP.

To highlight the advantage of FedParking, we compare our approach with the baseline approaches. We evaluate the training loss by using MSE. The performance regarding the MSE loss for three approaches are assessed in Fig.~\ref{s1}. FedParking ensures that knowledge of the local models is securely shared among the PLOs and simultaneously apply the LSTM model for accurate parking space estimation. As a consequence, FedParking converges significantly faster than the baseline approaches. For each PLO, experimental results show that compared with the baseline approaches, the proposed FedParking always achieves smaller MSE, thereby achieving a higher prediction accuracy.

\begin{figure}[t!]
	\centering
	\includegraphics[width=0.48\textwidth]{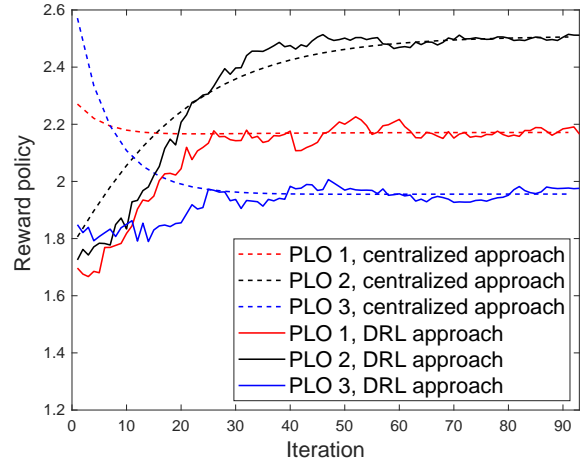}
	\caption{Performance comparison among different approaches without the parking capacity constraints.}
	\label{s3}
\end{figure}

\begin{figure*}[!htbp]
	\setlength{\belowcaptionskip}{-0.3 cm}
	\centering
	\subfigure[Case 1.]{
		\label{s4a} 
		\begin{minipage}[!htbp]{0.3\linewidth}
			\centering
			\includegraphics[width=1.0\textwidth]{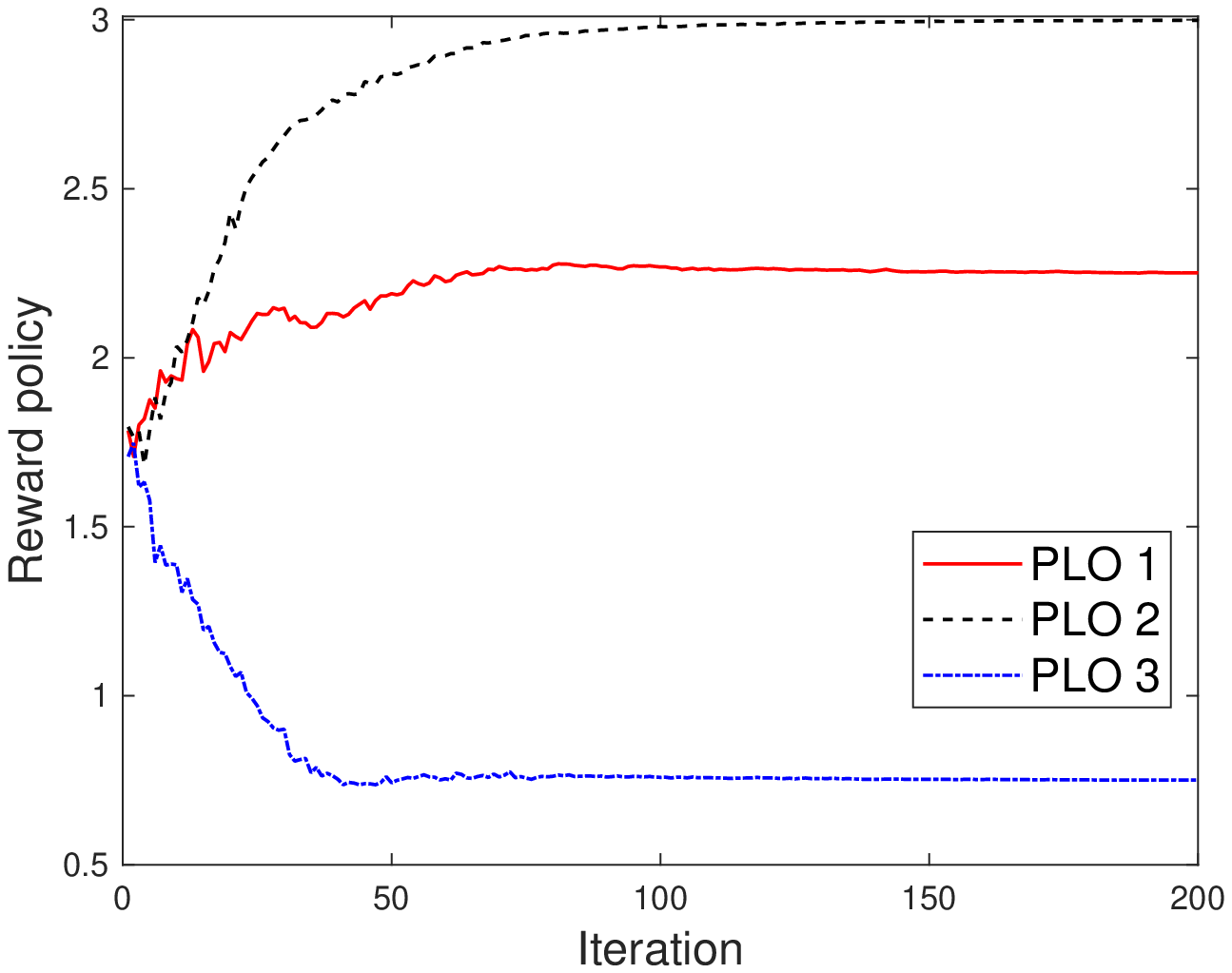}
	\end{minipage}}
	\subfigure[Case 2.]{
		\label{s4b} 
		\begin{minipage}[!htbp]{0.3\linewidth}
			\centering
			\includegraphics[width=1.0\textwidth]{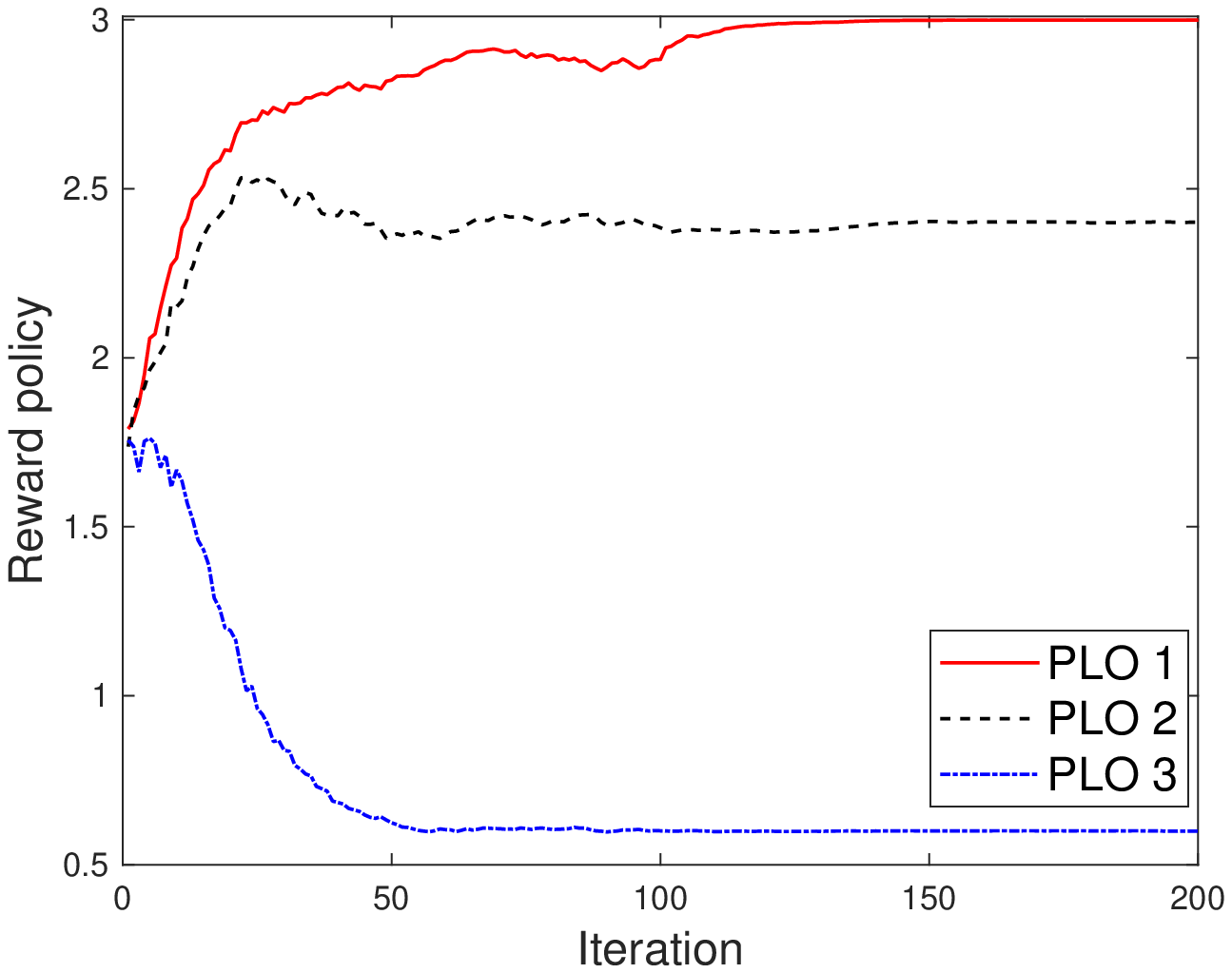}
	\end{minipage}}
	\subfigure[Case 3.]{
		\begin{minipage}[!htbp]{0.3\linewidth}
			\centering
			\label{s4c} 
			\includegraphics[width=1.0\textwidth]{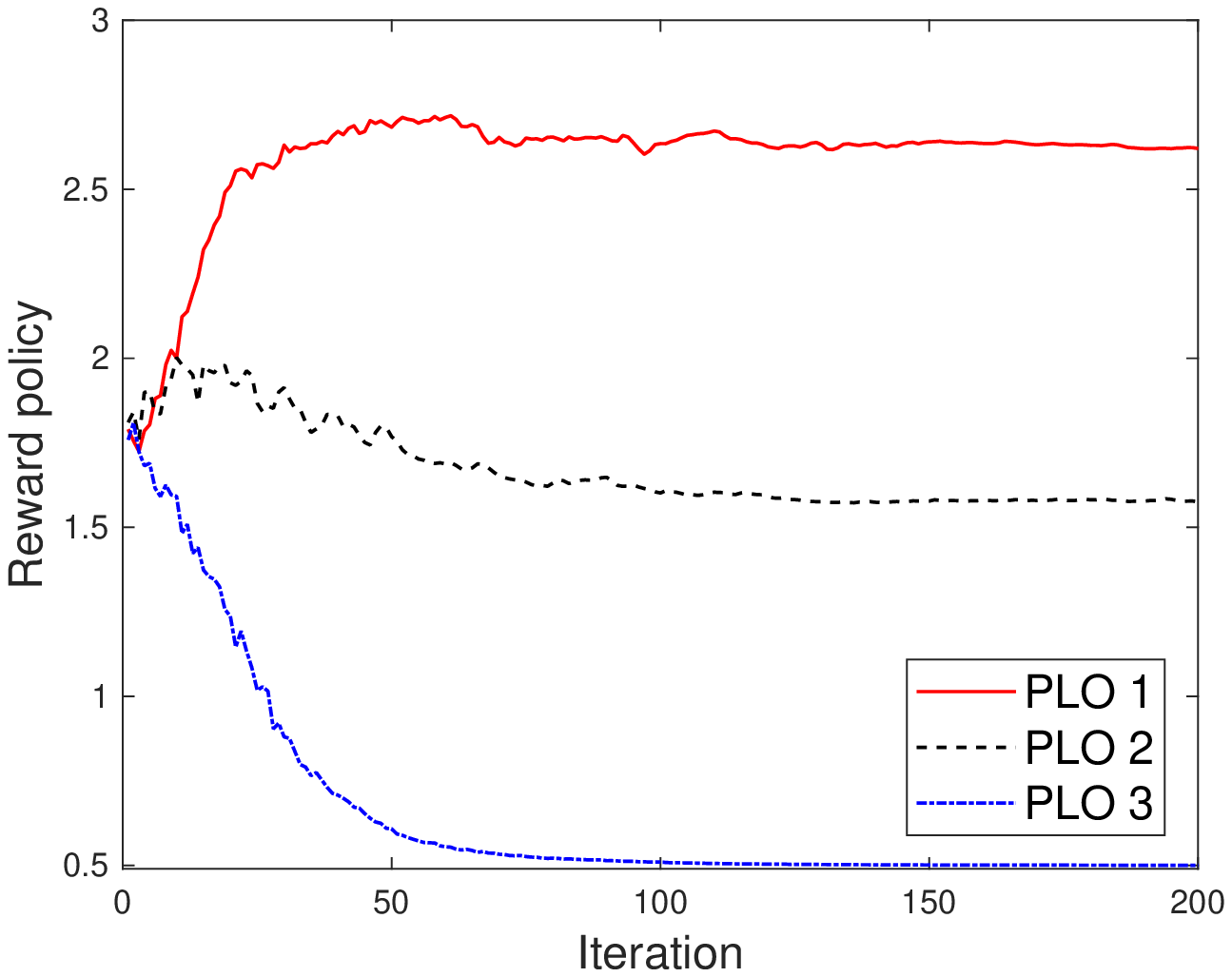}
	\end{minipage}}
	\caption{Convergence of the DRL approach with different parking capacity constraints.}
	\label{s4}
\end{figure*}

\begin{figure*}
	\setlength{\belowcaptionskip}{-0.3 cm}
	\centering
	\subfigure[Case 1.]{
		\label{s5a} 
		\begin{minipage}[!htbp]{0.3\linewidth}
			\centering
			\includegraphics[width=1.0\textwidth]{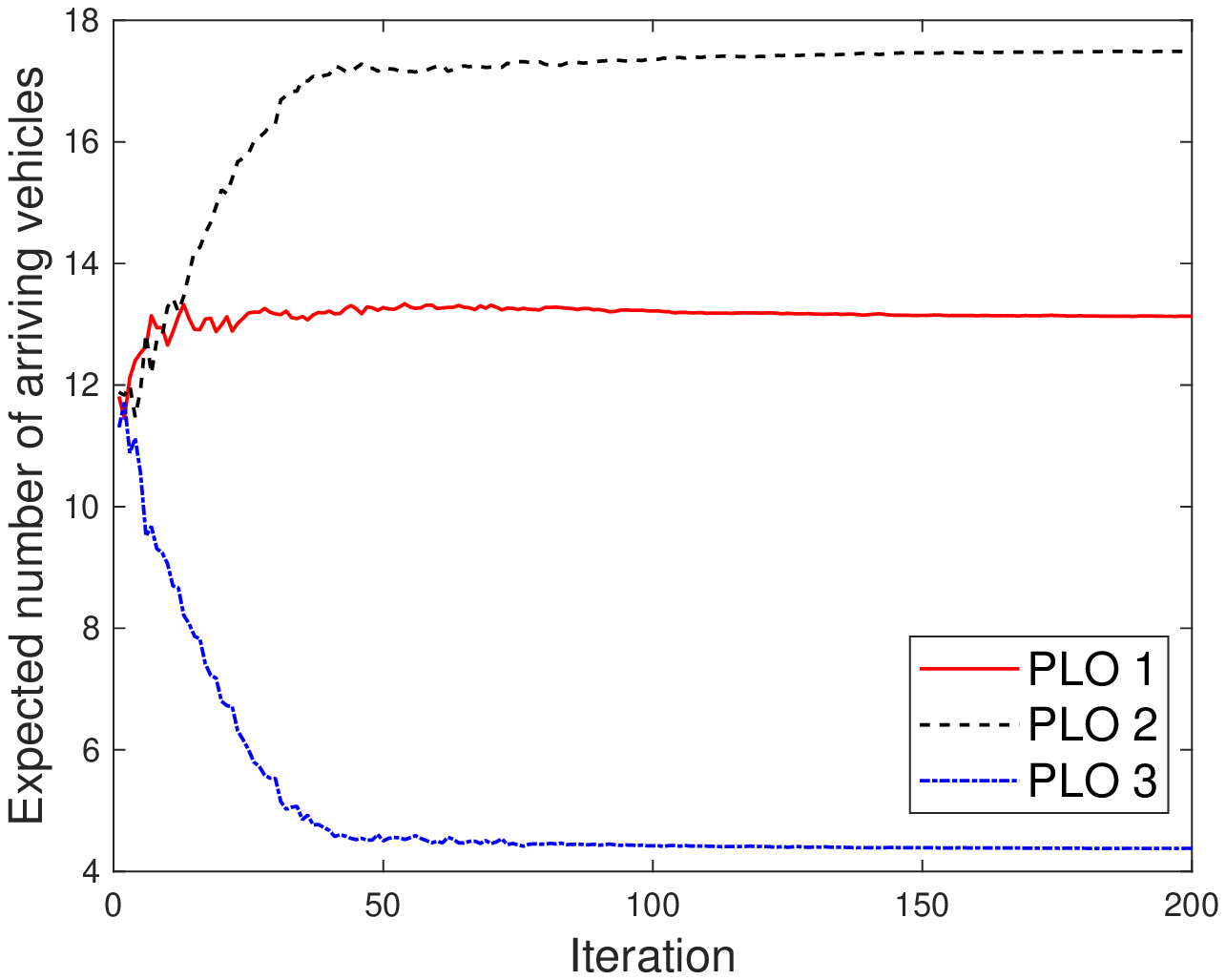}
	\end{minipage}}
	\subfigure[Case 2.]{
		\label{s5b} 
		\begin{minipage}[!htbp]{0.3\linewidth}
			\centering
			\includegraphics[width=1.0\textwidth]{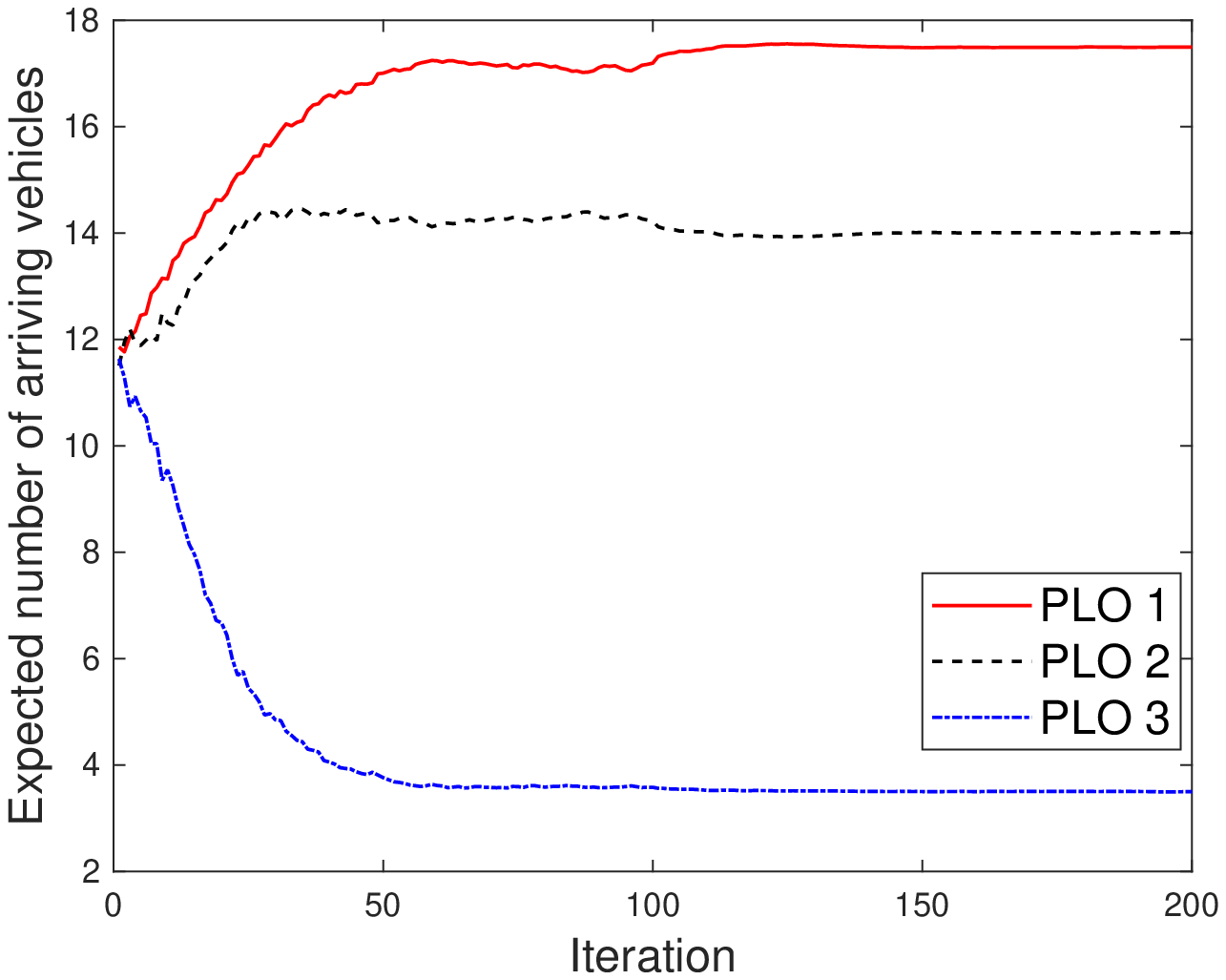}
	\end{minipage}}
	\subfigure[Case 3.]{
		\begin{minipage}[!htbp]{0.3\linewidth}
			\centering
			\label{s5c} 
			\includegraphics[width=1.0\textwidth]{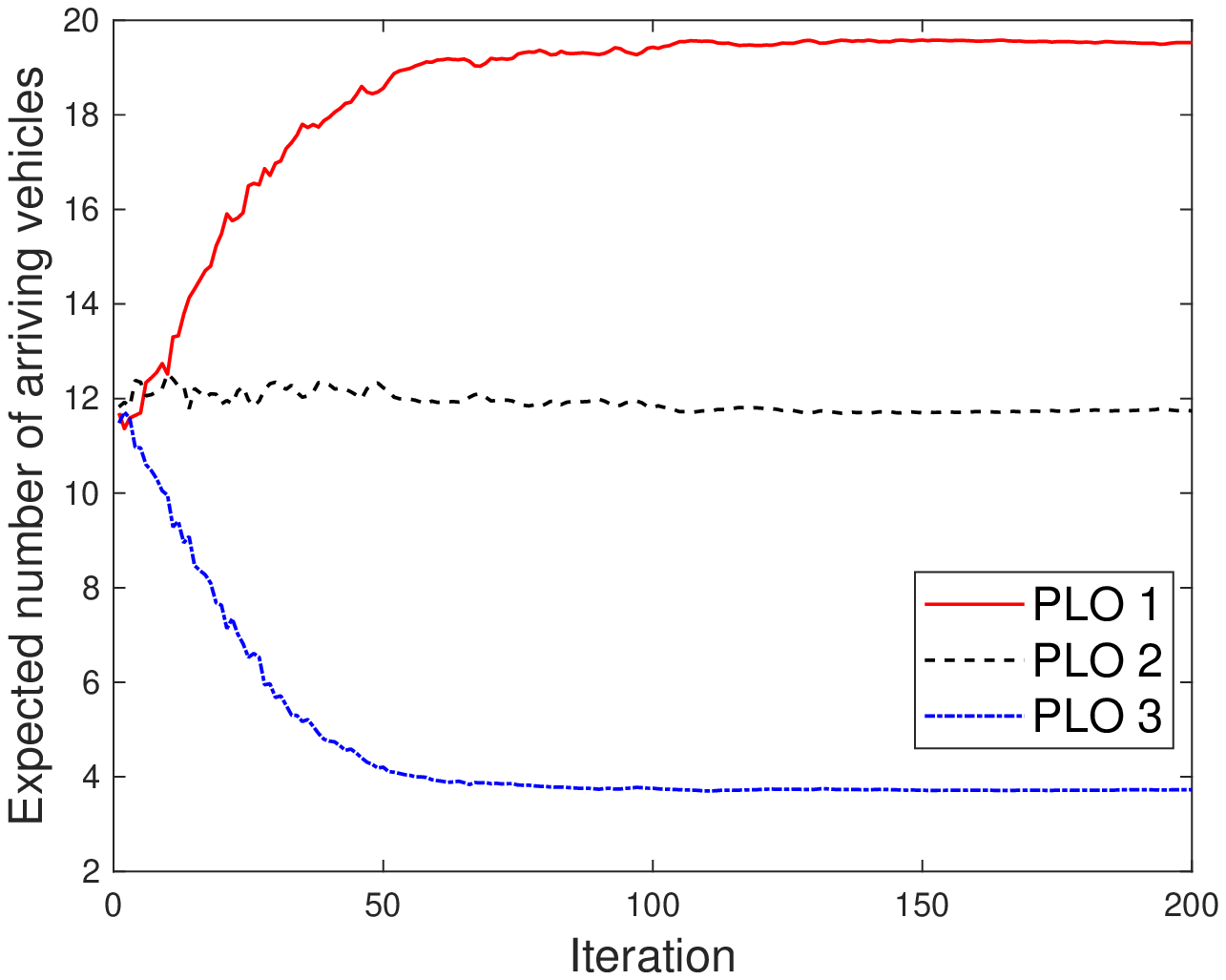}
	\end{minipage}}
	\caption{Comparison of the expected number of arriving vehicles under the DRL approach with different parking capacity constraints.}
	\label{s5}
\end{figure*}

\begin{figure}[t!]
	\centering
	\includegraphics[width=0.48\textwidth]{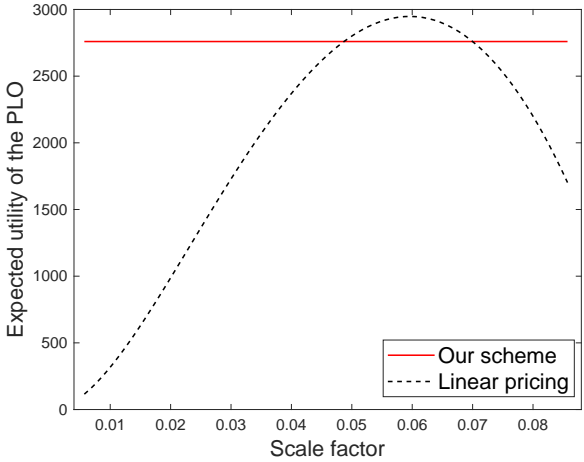}
	\caption{Comparison of the expected utility of a PLO under different schemes}
	\label{s51}
\end{figure}

\begin{figure*}[!htbp]
	\setlength{\belowcaptionskip}{-0.3 cm}
	\centering
	\subfigure[Comparison of the best response when varying $p$ and $r$.]{
		\label{s6a} 
		\begin{minipage}[!htbp]{0.3\linewidth}
			\centering
			\includegraphics[width=1.0\textwidth]{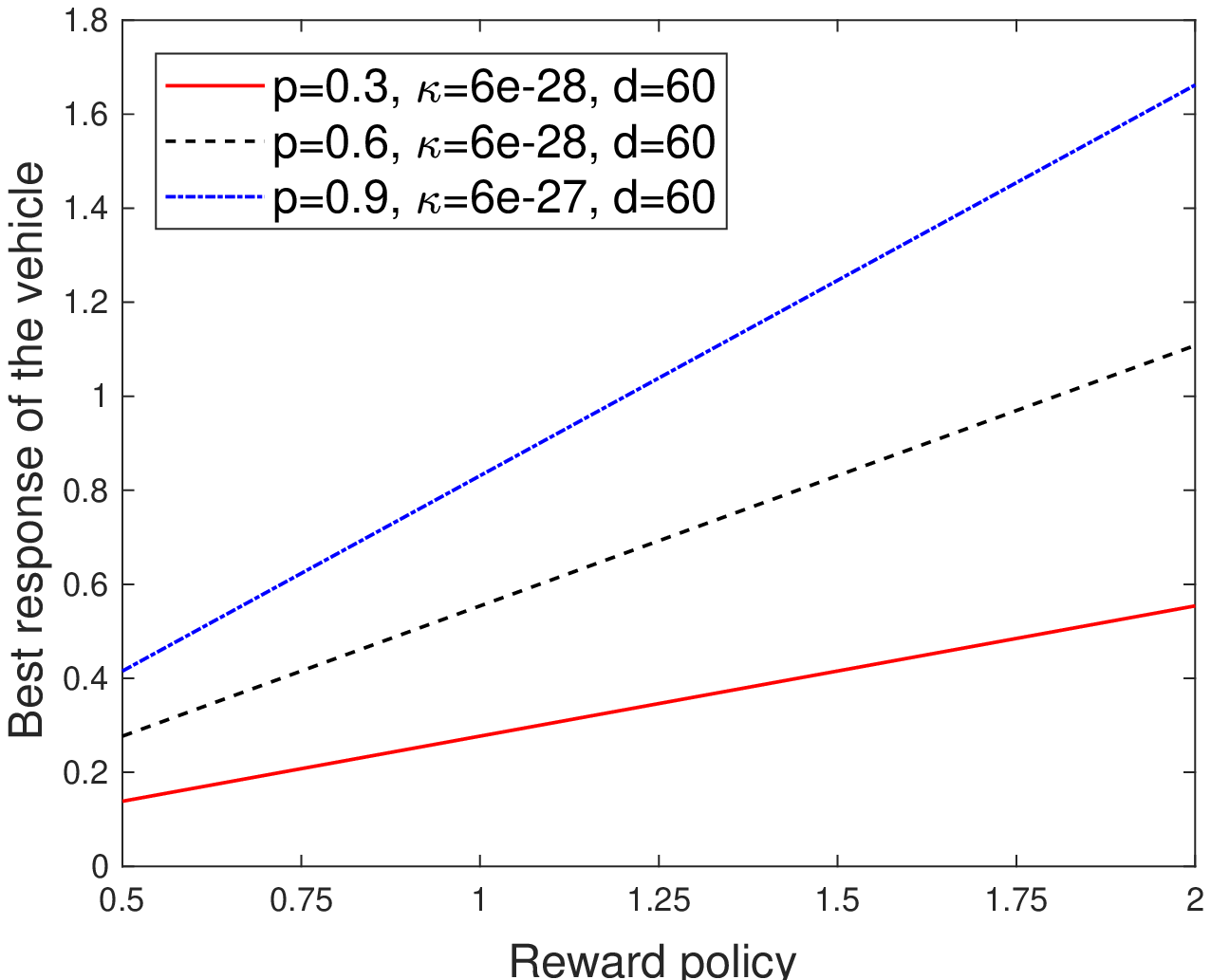}
	\end{minipage}}
	\subfigure[Comparison of the best response when varying $d$ and $r$.]{
		\label{s5b} 
		\begin{minipage}[!htbp]{0.3\linewidth}
			\centering
			\includegraphics[width=1.0\textwidth]{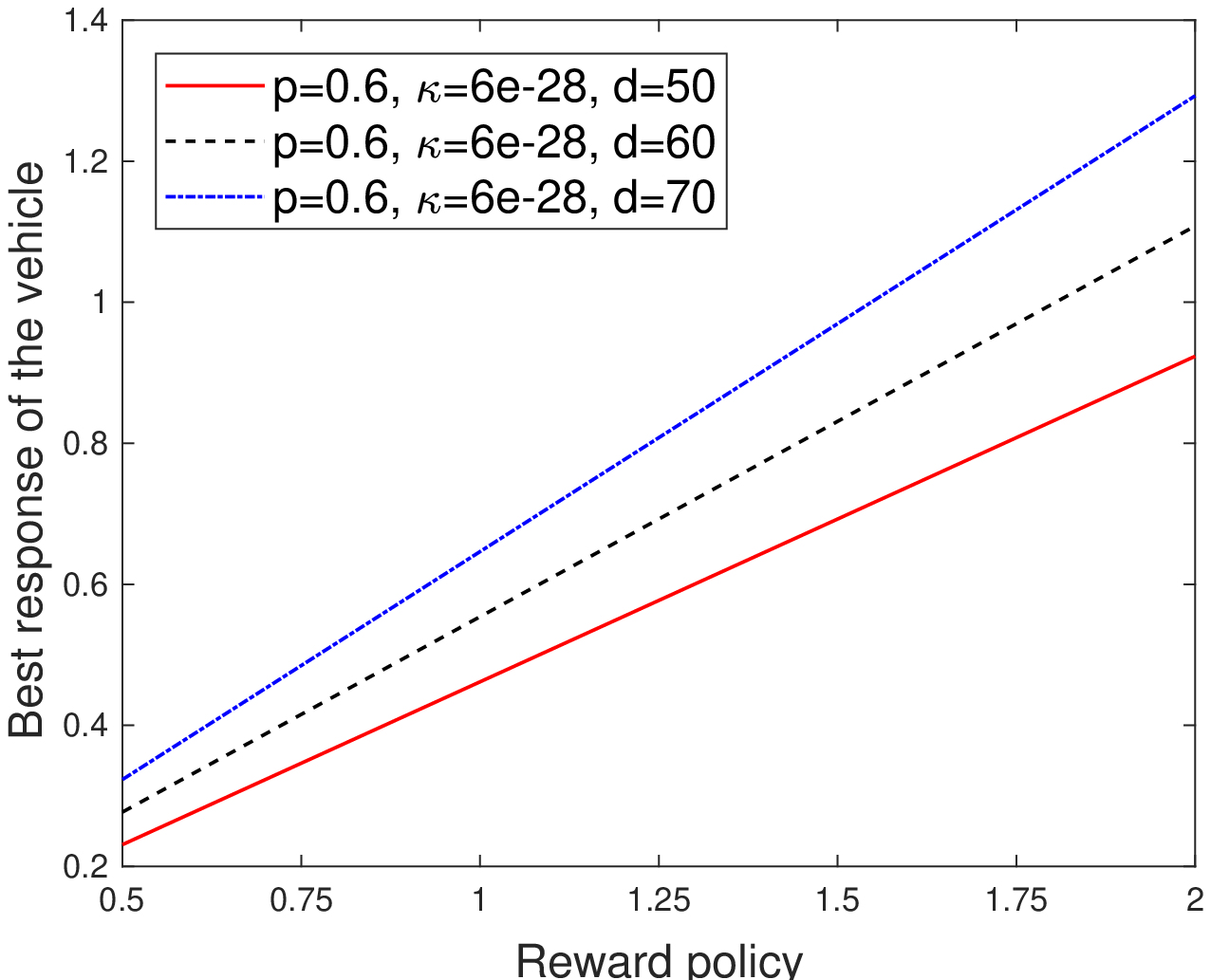}
	\end{minipage}}
	\subfigure[Comparison of the best response when varying $\kappa$ and $r$.]{
		\begin{minipage}[!htbp]{0.3\linewidth}
			\centering
			\label{s6c} 
			\includegraphics[width=1.0\textwidth]{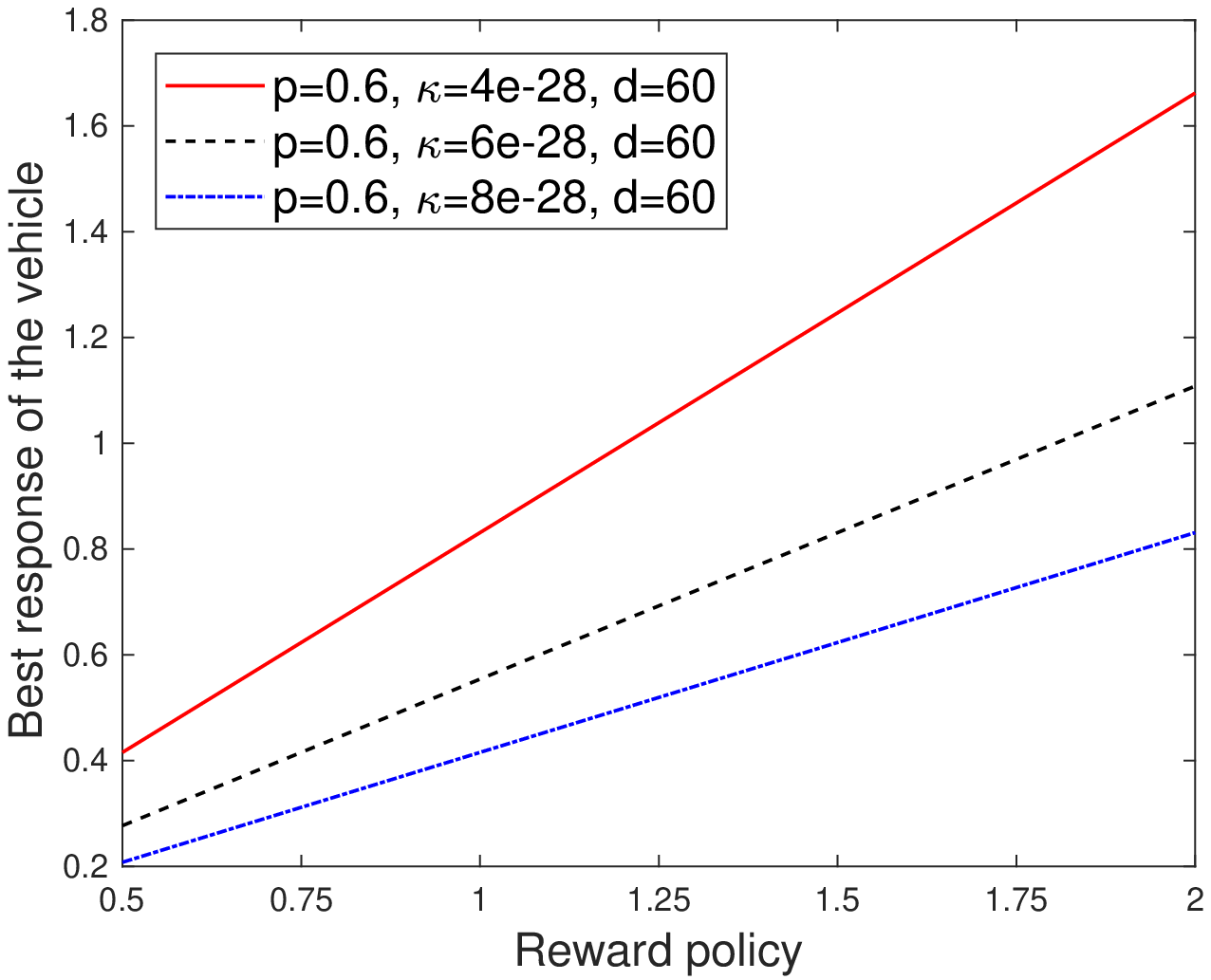}
	\end{minipage}}
	\caption{Comparison of the best response of a vehicle under different external and internal parameters.}
	\label{s6}
\end{figure*}

\subsection{DRL for Incentive Mechanism Design}
In the following, we perform numerical analysis of the DRL based incentive mechanism under different scenarios. We first investigate the performance of our DRL approach in the ideal case, where parking capacity constraints are neglected for the PLOs. To reach the Stackelberg equilibrium,  our work proposes a DRL approach to make each PLO as an agent learn a near-optimal solution in a distributed manner while the traditional work considers a centralized approach by collecting private information of all the PLOs and vehicles to execute the Jacobi based algorithm, as introduced in Section V-B. Here, we set the learning rate $\lambda=$1e-3 and constant $\Delta r=$1e-2 for the Jacobi based algorithm in the centralized approach. Without the parking capacity constraints, the penalty function is removed from reward function of each PLO in the DRL approach. As shown in Fig.~\ref{s3}, we compare the performance of our DRL approach with the centralized approach in the ideal case. The convergence of these two approaches is validated and the centralized approach converges relatively faster, owing to the perfect complete information condition. Beside, we observe that 
the learning error of the DRL approach is very minor. Our approach can finally converge to a near-optimal solution which can closely approximate the optimal solution in the centralized approach within the limited iterations. 

In addition, we investigate the performance of our DRL approach under different parking capacity constraints. For simplicity, we consider three cases of the distribution of the parking capacity constraints $\left[n^1,n^2, n^3\right]$, i.e., Case 1: $\left[15, 20, 5\right]$,  Case 2: $\left[n^1,n^2, n^3\right]=\left[25, 20, 5\right]$ and  Case 3: $\left[n^1,n^2, n^3\right]=\left[35, 20, 5\right]$ for 3 PLOs, respectively. From Fig.~\ref{s4}, we realize that the DRL approach converges finally under the parking capacity constraints. The expected number of the vehicles that choose to enter a specific parking lot is also controlled well  by adopting an appropriate reward policy, as illustrated by Fig.~\ref{s5}. For a PLO, reward policy $r$ referrings to monetary rewards per unit computing resource and unit time, roughly decreases with the reduction of the parking capacity constraint. Given a lower value of $n^j$, PLO $j$ is forced to provide less incentive rewards for arriving vehicles in order to  avoid
congestion to the parking lot. For example, when PLO 1 has only 15 free parking spaces in Case 1, $r^1$ is mainly influenced by the current parking capacity constraint. At the equilibrium point, the equality constraint $\sum\nolimits_{1 \le i \le I} {\rho _i^1}  = n^1$ is approximately satisfied. The similar observation is found for PLO 3, which has only 5 free parking spaces.  In Case 2, $n^1$is increased such that the PLO can improve $r^1$ on demand. At this time, PLO 1 has the maximal number of free parking spaces. To attract more vehicles, the PLO is encouraged to improve $r^1$, which is close to the upper limit in Case 2. By knowing the obvious improvement of $r^1$, PLO 2 and PLO 3 realize that more vehicles will be attracted by PLO 1. It is an alternative for them to improve their expected utilities by reducing $r^2$ and $r^3$ to save unnecessary monetary cost. Particularly, $r^2$ sharply decreases from 3.000 to 2.402 in Case 2. With the continuing increase of $n^1$, this leads to the further reduction of $r^2$ and $r^3$ in Case 3. The competitive strength of PLO 1 is enhanced when $n^1$ is  far greater than $n^2$ and $n^3$. To effectively improve the expected utility, PLO $1$ is active to reduce the addition monetary cost since $r^1$ with a slightly lower value still can attract sufficient vehicles, because of the simultaneous reduction of $r^2$ and $r^3$.

We compare the expected utility of a PLO under linear pricing scheme and our Stackleberg game based scheme in Fig.~\ref{s51}.  Here, we take PLO 1 as an example and set the parking capacity constraints in Case 3. According to the number of available parking spaces $n^1$ , it could be a simplified approach for PLO 1 to decide the reward policy $r^1$  by considering the linear relationship between $n^1$  and $r^1$ , namely,  ${r^1} = \zeta {n^1}$, where  $\zeta  \in [{r^{\min }}/{n^1},{r^{\max }}/{n^1}]$ is a scale factor. Compared with the linear pricing scheme, our scheme derives the optimal reward policy by the Stackelberg game model with DRL approach. With respect to the varying $\zeta$, the expected utility of PLO 1 is changing. Note that PLO 1 can adopt a fixed or dynamic reward policy by controlling the scale factor. But the determination of  $\zeta$ is a challenging task and cannot guarantee the high-level expected utility for the PLO all the time. By the traversal method, we seek the optimal value of  $\zeta$  to achieve the maximal expected utility. We find that the expected utility of PLO 1 calculated by our scheme significantly approximates the above maximal one. The approximation error is limited within about 6\%. Moreover, the linear pricing scheme is not always feasible since the baseline scheme neglects the parking capacity constraint. In this regard, our scheme applies the Stackelberg game theoretic approach and presents a learning based solution to quickly seek the suboptimal reward policy with the given parking capacity constraint.

Fig.~\ref{s6} shows the best response of a PV under the external and internal impacts. For the PV, the best response is in terms of the optimal number of computing resources shared to a PLO. Supposing that the PV is stimulated by PLO 1 and we study the best response $f^*$. The external impact refers to the given reward policy $r$, and internal factors include its preference parameter $p$, parking time $d$ and hardware parameter $\kappa$.   Clearly, with the increase of $r$, $f^{*}$ is gradually improved. According to Eqn.~(\ref{optf}), both $p$ and $d$ have a positive effect on the improvement of $f^{*}$ while $kappa$ has a negative effect. The results in Fig.~\ref{s6} are consistent with the above analysis. For example, when $d$ is increased from 50 to 70 minutes, the vehicle would like to share more idle computing resources and earn monetary rewards during the parked time, thereby $f^*$ increases about 40\%  with respect to  $r=1.5, p=0.6$ and $\kappa=6e-28$.  

In summary, we present a joint federated learning and DRL approach to design FedParking and ultimately facilitate the parking management in smart cities.  In FedParking,  a secure and distributed learning approach is applied among the PLOs for high-accuracy parking space estimation without requiring data transfer among them. The DRL approach resorts to the iterative and distributed decision making on the PLO side and may scarify the convergence speed compared with the centralized approach. But our approach guarantees the convergence accuracy, and does not require to collect private information of all the players of the Stackelberg game.

\section{Conclusion}
In this paper, we introduced FedParking to study the federated learning based parking space estimation with PVEC management. In the scheme,  PLOs were instructed to train a shared LSTM model for parking space estimation without exchanging the raw data among them. A parking space constraint was accordingly presented to each PLO, which acts as an incentive designer to determine how to stimulate the vehicles to enter the parking spaces and share their idle computing resources for offloading services in PVEC. We formulated the interactions among the PLOs and vehicles as a multi-leader multi-follower Stackelberg game and provided the theoretical Stackelberg equilibrium analysis under complete information. Considering dynamic arrivals of the vehicles and time-varying parking capacity constraints, a DRL approach was particularly proposed to reach the Stackelberg equilibrium in a distributed yet privacy-friendly manner. Finally, numerical results demonstrated that our scheme is effective and efficient for high-accuracy parking space estimation, and can seek a near-optimal solution under the complicated conditions.

\bibliographystyle{IEEEtran}
\bibliography{myreference}

\end{document}